\documentclass{article}

\usepackage[eandd, final]{neurips_2026}

\usepackage[utf8]{inputenc} 
\usepackage[T1]{fontenc}    
\usepackage{hyperref}       
\usepackage{graphicx}
\usepackage{url}            
\usepackage{booktabs}       
\usepackage{amsfonts}       
\usepackage{amsmath}
\usepackage{amssymb}
\usepackage{nicefrac}       
\usepackage{microtype}      
\usepackage{xcolor}         
\usepackage{makecell}
\usepackage{caption}
\usepackage{enumitem}
\usepackage{placeins}
\usepackage[most]{tcolorbox}
\usepackage{subcaption}
\usepackage{wrapfig}

\ifdefined\lessonbox
  \renewtcolorbox{lessonbox}[1][]{%
    enhanced, breakable,
    colback=white, colframe=black, boxrule=0.4pt, arc=2pt,
    boxsep=0pt,
    left=1em, right=1em,
    top=0.6ex, bottom=0.6ex,          
    before skip=6pt, after skip=10pt, 
    before upper=\parindent0pt\noindent,
    #1 
  }
\else
  \newtcolorbox{lessonbox}[1][]{%
    enhanced, breakable,
    colback=white, colframe=black, boxrule=0.4pt, arc=2pt,
    boxsep=0pt,
    left=1em, right=1em,
    top=0.6ex, bottom=0.6ex,
    before skip=6pt, after skip=10pt,
    before upper=\parindent0pt\noindent,
    #1
  }
\fi

\makeatletter
\ifdefined\variantbox
  \renewtcolorbox{variantbox}[2][]{%
    enhanced,
    breakable,
    colback=black!1,
    colframe=black!40,
    boxrule=0.3pt,
    arc=1.5pt,
    left=0.8em,right=0.8em,top=0.6ex,bottom=0.6ex,
    title={#2},
    fonttitle=\bfseries,
    boxed title style={
      colback=black!6, colframe=black!40,
      boxrule=0.3pt, arc=1.5pt,
      left=0.5em, right=0.5em, top=0.1ex, bottom=0.1ex
    },
    #1
  }
\else
  \newtcolorbox{variantbox}[2][]{%
    enhanced,
    breakable,
    colback=black!1,
    colframe=black!40,
    boxrule=0.3pt,
    arc=1.5pt,
    left=0.8em,right=0.8em,top=0.6ex,bottom=0.6ex,
    title={#2},
    fonttitle=\bfseries,
    boxed title style={
      colback=black!6, colframe=black!40,
      boxrule=0.3pt, arc=1.5pt,
      left=0.5em, right=0.5em, top=0.1ex, bottom=0.1ex
    },
    #1
  }
\fi
\makeatother

\ifdefined\gridbox
  \renewtcblisting{gridbox}{%
    enhanced, breakable, listing only, listing engine=listings,
    colback=black!3, colframe=black!60, boxrule=0.3pt, arc=1pt,
    boxsep=0pt, left=0.4em, right=0.4em, top=0.2ex, bottom=0.2ex,
    listing options={
      basicstyle=\ttfamily\scriptsize,
      columns=fullflexible, keepspaces=true,
      showstringspaces=false, aboveskip=0pt, belowskip=0pt
    }
  }
\else
  \newtcblisting{gridbox}{%
    enhanced, breakable, listing only, listing engine=listings,
    colback=black!3, colframe=black!60, boxrule=0.3pt, arc=1pt,
    boxsep=0pt, left=0.4em, right=0.4em, top=0.2ex, bottom=0.2ex,
    listing options={
      basicstyle=\ttfamily\scriptsize,
      columns=fullflexible, keepspaces=true,
      showstringspaces=false, aboveskip=0pt, belowskip=0pt
    }
  }
\fi

\newcommand{\ci}[2]{{\scriptsize [#1, #2]}}
\title{PotARCin: Multi-Dimensional Evaluation of Skill Acquisition in Abstract Reasoning Tasks}

\author{%
  Claas Beger \\
  Santa Fe Institute \\
  \texttt{claasbeger@santafe.edu} \\
  \And
  Ryan Yi \\
  Santa Fe Institute \\
  \texttt{ryi@santafe.edu} \\
  \And
  Melanie Mitchell \\
  Santa Fe Institute \\
  \texttt{mm@santafe.edu}
}

\begin{document}

\maketitle

\begin{abstract}
The Abstraction and Reasoning Corpus (ARC) has become a prominent benchmark for evaluating general abstract reasoning and fluid intelligence in AI models. Yet standard ARC evaluation considers only a single capability: producing the correct output grid for a test input. We argue that this narrow format fails to evaluate the diversity of abilities that genuine abstract skill acquisition should enable. We introduce PotARCin, a benchmark that extends ARC by assessing understanding of a task's underlying abstract rule across five dimensions: Definition, Classification, Constrained Generation, Editing, and Inversion. PotARCin employs programmatic methods to generate new task instances and transform given inputs for a given ARC task, enabling dynamic generative sampling beyond fixed input-output pairs. Across five state-of-the-art models evaluated on the ARC-AGI-1 training set, we observe a 25--52 percentage-point performance gap between standard ARC evaluation and evaluation on PotARCin, and find that multi-dimensional evaluation reorders models that standard accuracy ranks alike. We further investigate effects of generative sampling, difficulty of corruption types, and questions of self-consistency, showing that models frequently contradict their own formalized rule even where they have stated it correctly. We also introduce P-ARC, a held-out hand-crafted test set, on which models achieve 1--8\% accuracy across all five dimensions, underscoring the importance of more holistic evaluations of abstract reasoning capabilities.
\end{abstract}

\section{Introduction}
The Abstraction and Reasoning Corpus (ARC) \citep{chollet2019measureintelligence} has become a central benchmark for abstract reasoning and fluid intelligence in modern AI models. Given only a few demonstrations of grid transformations, ARC requires a solver to infer an underlying transformation rule and apply it to a novel test input grid to produce the corresponding output grid. For several years, ARC-AGI-1~\citep{ARC-AGI-1} posed a substantial challenge for state-of-the-art models. However, recent progress on AI reasoning models has led to the near-saturation of ARC-AGI-1 evaluation, motivating the release of follow-up benchmarks such as ARC-AGI-2~\citep{chollet2026arcagi2newchallengefrontier} and ARC-AGI-3~\citep{foundation2026arcagi3newchallengefrontier}. In this work, we nevertheless focus primarily on ARC-AGI-1. Rather than proposing harder ARC-style tasks, we ask whether high performance on the original output-generation task reflects the robust competence the benchmark is intended to measure.

We use ``skill acquisition'' in the sense of \citet{chollet2019measureintelligence}: intelligence as the efficiency with which a system acquires a new skill from limited task-specific experience. In ARC, the demonstrations \emph{are} that experience and the skill is the transformation rule inferred from them, so the term refers to this within-task inference process, not to longitudinal training or continual learning. PotARCin asks whether the inferred rule has been acquired as a \emph{reusable} skill across closely related applications.

\subsection{Background and Related Work}
\textbf{The Understanding Gap.} A growing body of work argues that high benchmark accuracy does not necessarily imply robust understanding of the underlying concept, rule, or skill a benchmark supposedly tests~\citep{ribeiro-etal-2020-beyond, mineault2026cognitivedarkmattermeasuring}. This concern is especially important for ARC, where the standard evaluation asks only whether a model produces the correct output grid. Recent work on an ARC-like benchmark called ``ConceptARC''~\citep{moskvichev2023conceptarc} shows that output accuracy can obscure whether models infer the intended abstractions, or whether they instead rely on unintended shortcuts or identify plausible rules they fail to execute~\citep{beger2026aimodelsperformhumanlike}.
A complementary perspective comes from the notion of ``Potemkin understanding'' introduced by \citet{pmlr-v267-mancoridis25a}. They argue that benchmark success only supports claims about conceptual understanding under the implicit assumption that when models ``misunderstand'' concepts, they misunderstand in ways similar to humans. When this assumption fails, a model may answer benchmark questions correctly while still failing closely related probes of the same concept---probes that a human who understood the original question would be expected to answer correctly. Their benchmark operationalizes this idea by testing whether models that can define a concept can also use it in closely related classification, constrained generation, and editing tasks~\citep{pmlr-v267-mancoridis25a}. We translate this multi-dimensional task framework to the ARC domain, replacing natural-language concepts with task-specific abstract transformation rules.

\textbf{Generative and Multi-Dimensional Evaluation.} These concerns are also aligned with broader critiques of evaluations using aggregate accuracy on static, held-out benchmarks. CheckList, for example, proposes behavioral testing through probes specific to a particular capability or test type rather than an undifferentiated measure of overall performance~\citep{ribeiro-etal-2020-beyond}; BIG-bench evaluates models over highly heterogeneous task families~\citep{srivastava2023imitationgamequantifyingextrapolating}; and GSM-Symbolic uses symbolic templates to generate controlled variants of mathematical reasoning problems, revealing brittleness to small variations in problem instantiation~\citep{mirzadeh2025gsmsymbolicunderstandinglimitationsmathematical}. Other recent benchmarks similarly move toward multi-dimensional evaluation of agent behavior, for example by measuring software-engineering performance through more fine-grained capabilities such as bug fixing, test generation, code-reviewing, and style fixing~\citep{sonwane2026omnicodebenchmarkevaluatingsoftware}. 

\textbf{ARC and Program Synthesis.} ARC has also been studied through the lens of program synthesis. Some methods attempt to solve ARC tasks by inducing programs that implement the inferred transformation rule~\citep{li2024combining, pourcel2026selfimprovinglanguagemodelsevolutionary}, while others use programs to generate procedural variants of ARC-like tasks~\citep{hodel2024addressingabstractionreasoningcorpus, moffitt2025arcgenmimeticproceduralbenchmark}. 

Together, these lines of work motivate a benchmark that is both generative and multi-dimensional. PotARCin brings this perspective to ARC by using explicit generator and verifier programs to define the space of valid input-output pairs and to sample novel instances beyond the fixed demonstrations. This generative structure allows us to evaluate whether a model has acquired the underlying transformation rule as a reusable abstraction rather than simply solved a single held-out input. We then test this rule-abstraction competence across five dimensions---definition, classification, constrained generation, editing, and inversion---in the style of \citet{pmlr-v267-mancoridis25a}.

\subsection{Motivation and Contributions}
Following this perspective, we view each ARC task as defining a small, task-specific skill. This skill is not merely the ability to map a particular input grid to its corresponding output grid, but the ability to infer and use the transformation rule underlying the task. Standard ARC evaluation tests one use of this skill, namely, forward transformation: given a new input grid, the model must produce the corresponding output grid. However, when a human has robustly acquired such a skill, we expect them to be able to use it in other closely related settings, such as formally defining the rule, repairing examples that violate it, or distinguishing valid from invalid input-output pairs. Failures on these related operations therefore reveal gaps in skill competence that output accuracy alone would not reveal. Such gaps are particularly important when considering the deployment of more complex skills in real-world settings, where failures may be less visible and more consequential.

PotARCin aims to measure these gaps, and is named to reflect the connection to \mbox{\citep{pmlr-v267-mancoridis25a}}. Inspired by the Potemkin Understanding framework, this benchmark is designed to expose cases in which a model's accuracy on ARC tasks gives the appearance of rule understanding without the corresponding underlying competence. Our \textbf{contributions} are:

\begin{itemize}[leftmargin=*, itemsep=1pt, topsep=1pt]
    \item \textbf{PotARCin}, a benchmark evaluating ARC rule competence across five dimensions, with five procedures for generating plausible corrupted pairs from task-specific generator and verifier programs.\footnote{Code released at \url{https://github.com/ClaasBeger/PotARCin}.}
    \item \textbf{P-ARC}, a held-out set of 50 hand-crafted ARC-style tasks with generators, verifiers, and human-generated corruptions, enabling evaluation outside the public ARC-AGI-1 distribution.
    \item An evaluation of \textbf{five frontier models} showing large drops from output-grid accuracy to multi-dimensional competence, including a ranking reversal in which the weakest model by output-grid accuracy is not the weakest by rule competence.
    \item Two \textbf{targeted diagnostics} separating the mechanical cost of conjoining five dimensions from genuine inconsistency: a matched-budget comparison of cross- versus within-dimensional probing, and a self-consistency analysis conditioned on a correct executable Definition.
\end{itemize}


\section{Methodology}\label{sec:methodology}

\textbf{Generators and Verifiers.} PotARCin relies on two task-specific program types: generators and verifiers. A \textit{generator} samples new input-output pairs for a given ARC task, thereby defining a distribution of valid task instances. A \textit{verifier} implements the task transformation by mapping a candidate input grid to its corresponding output grid, providing an automated way to evaluate whether proposed outputs satisfy the underlying transformation rule.

Two prior works provide generator programs for ARC-AGI-1 tasks. RE-ARC~\citep{hodel2024addressingabstractionreasoningcorpus} introduces a domain-specific language for constructing functions that can generate diverse new input-output pairs. ARC-GEN~\citep{moffitt2025arcgenmimeticproceduralbenchmark}, by contrast, emphasizes what the authors call ``mimetic similarity'' in the design of its generators, requiring generated examples to closely preserve the constraints and distributional properties of the original demonstrations. This distinction is important for our setting: a generator may be logically consistent with the original demonstrations, yet shift or broaden the rule enough that a solver could not reasonably infer it from the demonstrations alone. For PotARCin, we therefore require generated examples to be not only valid under a task rule, but also sufficiently mimetic with respect to the original demonstrations.

\begin{figure*}[htbp]
  \centering
  \includegraphics[width=0.7\textwidth]{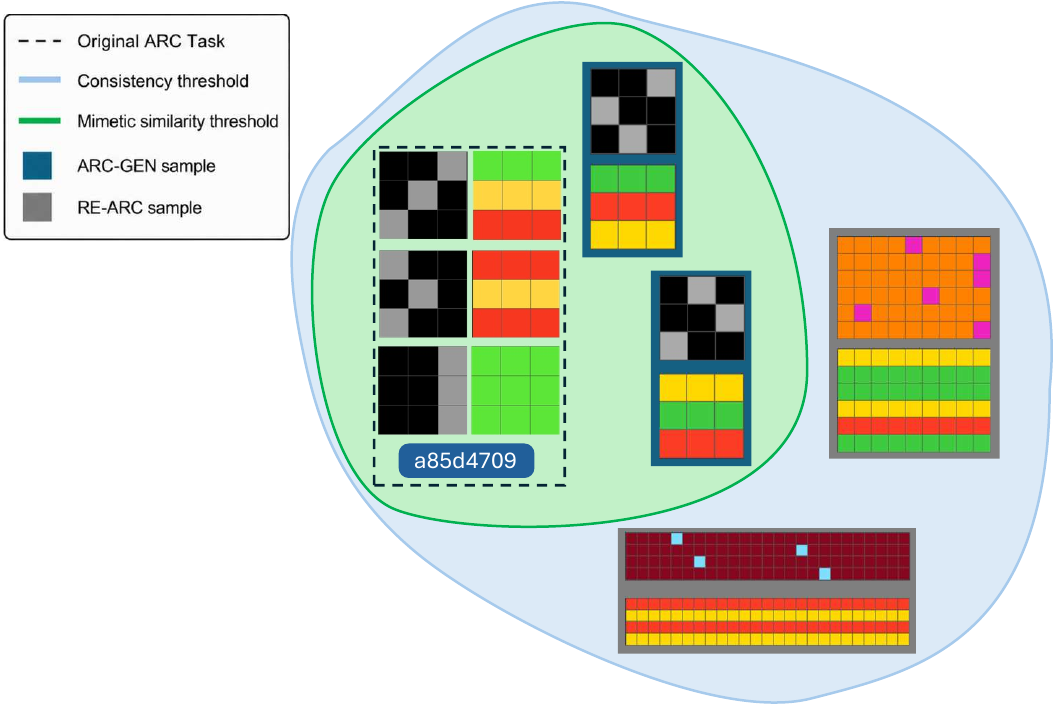}
  \caption{Consistency and mimetic similarity boundaries for ARC-AGI-1 task \texttt{a85d4709}.}
  \label{fig:sim_circles}
\end{figure*}

To illustrate, consider ARC-AGI-1 task \texttt{a85d4709}, whose demonstrations are shown within the dotted outline in \autoref{fig:sim_circles}. While ARC-GEN preserves the task's demonstrated $3 \times 3$ structure with one gray cell per row, in which gray-cell position determines row color, RE-ARC instead generalizes to variable-width grids by dividing rows into three regions and recoloring according to the region containing an outlier cell. This rule is consistent with the demonstrations but significantly changes the task distribution, and a solver would have little reason to infer it from the original task alone. We return to this distinction for the Definition dimension in \autoref{sec: Results}.

In addition to generators, PotARCin requires task-specific verifiers, programs that, given an input grid, generate the output grid according to the ground-truth rule governing the task. RE-ARC provides verifiers for the ARC-AGI-1 training corpus~\citep{hodel2024addressingabstractionreasoningcorpus}. ARC-GEN does not provide verifiers directly, but its associated Kaggle code-golf competition collected short verifier programs for ARC-AGI-1 training tasks~\citep{google-code-golf-2025}. We use the first-, third-, and fifth-place entries alongside the RE-ARC verifiers.

To ensure correctness, we validate candidate verifiers against multiple sources of task instances: the original ARC-AGI-1 train and test examples, the ARC-GEN stable dataset (about 250 pairs per task), and 50 random samples from the corresponding generators. Verifiers count as valid only if they produce the expected output grid for all tested examples. Across all available verifier sources, we obtain at least one valid verifier for 394 of the 400 tasks in the ARC-AGI-1 training set; we manually implemented verifiers for the remaining six tasks and validated them using the same process.

\textbf{Evaluation Dimensions.} Using these generator and verifier programs, we define five evaluation dimensions for each ARC task. Four of these dimensions are modified versions of the framework from \citet{pmlr-v267-mancoridis25a}; the fifth, inversion, is specific to the ARC domain. Given the demonstrations for an ARC task, a model is evaluated along the following five dimensions:

\begin{itemize}[leftmargin=*, itemsep=2pt, topsep=2pt]
    \item \textbf{Definition:} The model must write a general-purpose Python program that implements the inferred transformation rule, analogous to a verifier. We measure the accuracy of this program on the task demonstrations, the original test input, the ARC-GEN stable dataset, and 50 samples from the underlying generator, where accuracy is defined as producing the correct output grid. Non-executable programs are rare and account for little of this dimension's error (\autoref{sec:definition_errors}).
    \item \textbf{Classification:} The model is shown five candidate input-output pairs (in addition to the original demonstrations) and must decide whether each pair follows the same transformation rule as the demonstrations. Candidate pairs include both valid pairs sampled from the task generator and invalid pairs produced using the corruption procedures described below. Classification is strict at the task level: all five candidate judgments must be correct to pass.
    \item \textbf{Constrained Generation:} The model must generate a novel input-output pair that demonstrates the transformation rule it has inferred from the demonstrations. We evaluate the proposed pair by applying the task verifier to the model's input grid and checking whether the verifier output matches the model's proposed output grid.
    \item \textbf{Editing:} The model is given a corrupted input-output pair and must repair it while remaining close to the corrupted pair under a normalized edit-distance constraint.\footnote{We use a normalized Hamming distance with padding. We align the corrupted grid and the model-predicted grid on a shared padded bounding box with height $h=\max(h_a,h_b)$ and width $w=\max(w_a,w_b)$, using fill value -1. The distance is the number of differing cells, normalized by $h \cdot w$. This distance is computed separately for the input and output grids; both must be at most $0.70$. A similar measure is applied to ensure distance from training examples for constrained generation.} The repaired pair is accepted only if it satisfies this constraint and the verifier maps the edited input to the edited output.
    \item \textbf{Inversion:} The model is given an output grid and must infer a possible input grid that would produce it under the ground-truth task rule. We evaluate the proposed input by applying the task verifier to it and checking whether the resulting output grid matches the provided output grid.
\end{itemize}

\begin{figure}[htbp]
  \centering
  \includegraphics[width=1\textwidth]{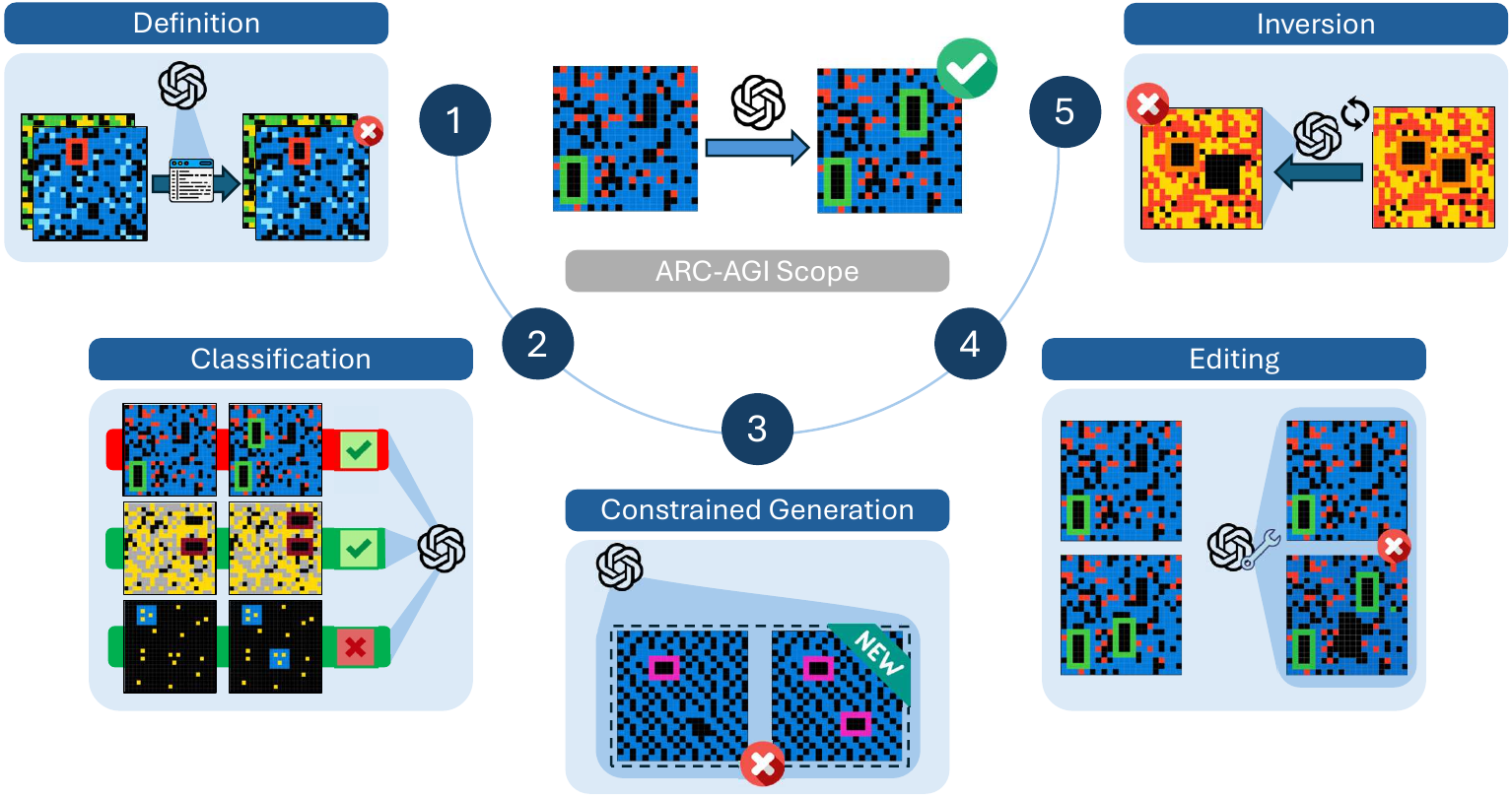}
  \caption{Overview of the PotARCin benchmark. The shown ARC task is \texttt{890034e9}, for which GPT-5.4 solved the standard test-grid task but failed all additional dimensions.}
  \label{fig:overview}
\end{figure}

Taken together, these five dimensions probe different uses of the same latent transformation rule, given the same demonstrations. Failures on them can reveal shallow, brittle, or unintended rule understanding that output-grid accuracy alone may not capture~\citep{beger2026aimodelsperformhumanlike, pmlr-v267-mancoridis25a, mineault2026cognitivedarkmattermeasuring}. Because all five are scored with explicit generators and verifiers, PotARCin broadens ARC evaluation while preserving fully automated scoring. One caveat applies throughout: the dimensions differ in \emph{verification burden}. Definition is checked against many sampled instances, while Inversion is checked against only one. We treat the many Definition samples as a single evaluation instance, used to confirm that the program captures the rule broadly rather than overfit to a narrow set of examples, but cross-dimension difficulty comparisons should be read with this asymmetry in mind.


\textbf{Corruption Types.} To evaluate Classification and Editing, PotARCin requires invalid input-output pairs that are still close enough to the task distribution to be challenging. We therefore construct corrupted pairs using five complementary procedures. We then apply the task verifier to ensure that the corrupted pair does not satisfy the target transformation rule. Examples of all corruption types are provided in \autoref{sec:corruption_types}.

\begin{itemize}[leftmargin=*, itemsep=2pt, topsep=2pt]
    \item \textbf{Instance Mismatch:} We pair an input grid with an output grid from a different valid instance of the same task. To make the mismatch plausible, we use an edit-distance heuristic to search the ARC-GEN stable dataset and the original training examples for outputs similar to the correct output, then sample from the top three ranked candidates.
    \item \textbf{Retrieval:} We retrieve input-output pairs from other ARC tasks with similar ``task embeddings''. To construct these embeddings, we adapt the V-ARC architecture~\citep{hu2025arcvisionproblem}, which combines a vision transformer backbone with an explicit task embedding vector intended to encode the underlying task rule. More specifically, we adopt the approach of \citet{deliège2026implicitruleinductiontesttime}, who freeze the transformer backbone and tune only the task embedding. We then build a vector database and use cosine similarity to retrieve pairs from related but distinct transformation rules.
    \item \textbf{H-ARC:} We use incorrect, human-generated output grids from the H-ARC dataset~\citep{LeGris2025}. As one of the largest-scale studies on human ARC performance, H-ARC provides a sizable dataset of human-generated answers, including many erroneous solutions. After filtering empty solutions, 374 ARC-AGI-1 training tasks have at least one incorrectly constructed output available, with an average of 8.36 per task. These errors provide naturally occurring invalid outputs that arise from human attempts to solve the same tasks.
    \item \textbf{Corrupted Verifier:} We generate corrupted outputs by mutating verifier programs. For RE-ARC verifiers, which are written as sequences of DSL assignments, we remove a small number of intermediate assignments and reconnect the remaining program. For other verifier implementations, we construct the abstract syntax tree and perform a binary-operation prune, replacing a binary expression with its left subtree. Applying these corrupted verifiers to valid inputs yields outputs produced by a perturbed version of the task rule.
    \item \textbf{Color Flip:} We locally perturb a valid generated pair by sampling a cell from either the input or output grid and reassigning the color of that cell, together with orthogonally connected cells of the same color, to a new color. To avoid trivial background changes, we avoid black cells.
\end{itemize}

At evaluation time, we sample from a fixed mixture over valid pairs and the five corruption types; the exact weights, which differ between datasets to reflect differences in corruption quality, are given in \autoref{tab:corruption_weights}. Because ARC tasks are generally underspecified, these labels are defined relative to the designated ground-truth rule instantiated by the task generator and verifier; a pair labeled as corrupted may still be compatible with some alternative rule that also explains the demonstrations. For Editing, we sample a single invalid pair from the latter three types.

\textbf{Evaluation Protocol.} In our evaluation, each of the 400 ARC-AGI-1 training tasks is evaluated over the five dimensions described above, as well as for the original output-grid generation skill. Each skill is evaluated independently, in a new context window.  ARC performance is commonly reported using pass@2, where a task is counted as solved if either of two submitted output grids exactly matches the ground truth~\citep{chollet2025arcprize2024technical}. We do not adopt pass@2 as our primary metric, both because it does not translate cleanly across all PotARCin dimensions and because our goal is to evaluate rule competence across multiple uses of the same inferred transformation. Instead, we report \textit{full-task accuracy}: the percentage of tasks for which a model passes all five PotARCin dimensions under a given sampling run.

Because full-task accuracy is a coverage metric over five dimensions, we report alongside it the \textit{independence baseline}: the product of the five marginal accuracies, i.e.\ the joint rate expected if per-dimension pass events were independent. All runs are repeated twice, and we report pooled Wilson $95\%$ confidence intervals over the $n{=}800$ (ARC-AGI-1) or $n{=}100$ (P-ARC) task evaluations rather than standard deviations over two runs, which are too few to estimate run-to-run variance and can suggest perfect stability when two runs happen to yield equal counts. Equal counts also hide churn in \emph{which} tasks pass: GPT-5.4 passes 185 and 188 ARC-AGI-1 tasks across its runs, intersecting on only 146 (Jaccard $0.64$). Full-task pass-set overlap ranges from $0.49$ (Kimi K2.5) to $0.79$ (Gemini 3.1 Pro) across models; per-model counts are in \autoref{tab:run_overlap}.

We evaluate five frontier models with strong ARC-AGI-1 performance: GPT-5.4, Gemini 3.1 Pro, Claude Opus 4.6, Kimi K2.5, and MiniMax M2.5. Because PotARCin multiplies the number of evaluations per ARC task across dimensions and samples, we generally use low reasoning-effort settings unless otherwise noted; Claude receives a 120K thinking budget, as in prior ARC evaluations. We use the public ARC-AGI-1 training set as a controlled and favorable evaluation setting, since it is accessible, widely studied, and likely easier than held-out ARC-AGI-1 splits~\citep{chollet2025arcprize2024technical,LeGris2025}. Results on ARC-AGI-1 training tasks should therefore be interpreted with caution: o3, GPT-5.4's predecessor, and Claude Opus 4.6 have been reported to have trained on ARC-AGI-1 training data~\citep{arcprizeOpenAIBreakthrough,Anthropic}. Other frontier models likely have had similar exposure. Even in this favorable setting, PotARCin exposes substantial drops relative to standard output-grid accuracy.

\textbf{P-ARC Test Set.} To estimate performance on unseen tasks, we construct a new held-out ARC-style test set, P-ARC, which we release at \url{https://huggingface.co/datasets/ClaasBeger/P-ARC}. P-ARC consists of 50 hand-crafted tasks, each with corresponding generator and verifier programs. Because H-ARC errors are unavailable for new tasks, each task instead has three human-made corruptions: erroneous outputs collected during the feasibility check described below, and errors designed by the task's creator. Our human protocol is a \emph{feasibility check} embedded in the design process, not a formal study with naive participants: each task was shown to two or three team members other than its author and retained only if at least one produced the correct output; solve times were not recorded. For every task, a member other than the author inspected at least 50 generator-produced examples and confirmed they realize the intended rule. Taxonomy and further calibration details are in \autoref{sec:parc_details}. Although difficult to quantify, we qualitatively estimate P-ARC to lie between ARC-AGI-1 and ARC-AGI-2 in difficulty.

\section{Results}\label{sec: Results}

\begin{figure*}[htbp]
  \centering
  \includegraphics[width=1\textwidth]{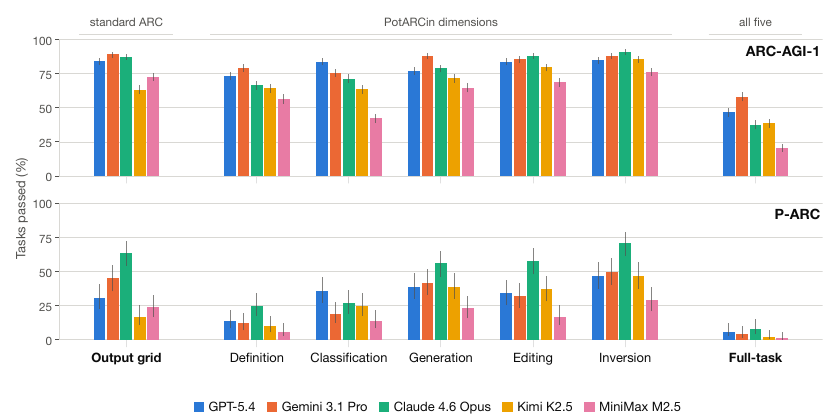}
  \caption{Performance across all seven measures for 400 ARC-AGI-1 training tasks and 50 P-ARC tasks, pooled over two runs ($n{=}800$ and $n{=}100$ task evaluations); whiskers are Wilson $95\%$ intervals. Both panels share a vertical scale. Exact values are in \autoref{tab:main_results}.}
  \label{fig:performance_overview}
\end{figure*}

We report three levels of performance in \autoref{fig:performance_overview}: standard output-grid correctness, dimension-specific accuracy, and full-task accuracy, for both the ARC-AGI-1 training set and the held-out P-ARC test set. Exact values and the independence baseline are reported in \autoref{tab:main_results}.

\textbf{ARC-AGI-1.} On the standard output-grid task, the proprietary models achieve 84--89\% accuracy on the ARC-AGI-1 training set, while MiniMax reaches 72.9\% and Kimi 63.4\%. Full-task accuracy, requiring success across all five PotARCin dimensions, is only 21--58\%, a drop of 25--52 percentage points relative to output-grid accuracy. Definition and Classification are consistently the hardest and Inversion the easiest, subject to the verification-burden asymmetry noted in \autoref{sec:methodology}.

The multi-dimensional evaluation also reveals differences between models that are largely hidden by output-grid accuracy alone. Gemini and Claude achieve similar performance on the standard ARC output-grid-generation, but Gemini outperforms Claude by 20.6 percentage points in full-task accuracy. More strikingly, Kimi K2.5 has the lowest output-grid accuracy of the five models (63.4\%) yet nearly doubles MiniMax's full-task accuracy (38.6\% versus 20.6\%), and matches Claude's (37.6\%, with overlapping intervals) despite trailing it by 24.2 points on the standard task. Across individual dimensions, proprietary models often stay closer to their output-grid accuracy. Mimetic similarity also matters for Definition: evaluated on the broader RE-ARC-generated distribution, Definition accuracy drops to roughly 3\%, supporting our concern that inferred programs do not generalize to shifts too wide to infer from the demonstrations.

\textbf{P-ARC.} On the held-out P-ARC test set, output-grid accuracy is lower overall and performance gaps between models are larger. Full-task accuracy falls to 1--8\% for every model, indicating the difficulty of acquiring robust multi-dimensional competence on unseen tasks. The ordering across dimensions matches ARC-AGI-1: Definition is hardest for all five models, at 6--25\%, while Inversion is easiest, at 29--71\%. Claude is strongest on five of the six reported measures (with GPT-5.4 leading on Classification), although this performance is accompanied by a marked increase in token consumption despite low reasoning effort (see \autoref{sec:adaptive_token}). Overall, PotARCin exposes substantial gaps in skill competence on newly constructed tasks outside the public ARC-AGI-1 training distribution.

As a stricter control, excluding exact matches to the ARC-GEN stable set lowers ARC-AGI-1 Constrained Generation accuracy by 3.4--7.6 points (\autoref{sec:cg_strict}).

\subsection{Isolating Cross-Dimensional Failures}\label{sec:gap}

Part of the drop is arithmetic, since full-task accuracy conjoins five dimensions. Observed full-task accuracy nevertheless \emph{exceeds} the independence baseline of \autoref{tab:main_results} for every model and both datasets, by $+7.4$ to $+18.5$ points on ARC-AGI-1, so per-dimension pass events are positively correlated rather than independent, as expected when tasks vary in difficulty: easy tasks tend to be passed across dimensions and hard ones failed across them. Marginal independence is therefore a lower bound on the joint rate. The analyses in this section isolate the component of multi-dimensional failure that repeated probing in a single format would not expose.

The sharper question is whether cross-dimensional probing surfaces failures that repeated probing within one dimension would miss. We test this at a \emph{matched evaluation budget} on the 20 ARC-AGI-1 tasks GPT-5.4 initially passed across all five dimensions, using ten same-seed repeats per task. Evaluating five different dimensions once exposes at least one failure in $25.5\%$ of trials, versus $20.2\%$ when a single typical dimension is sampled five times (all $\binom{10}{5}$ subsets; task-paired $t$-test $p{=}0.030$, Wilcoxon $p{=}0.028$, bootstrap $95\%$ CI on the difference $[+1.0, +10.5]$ pp). Two alternative aggregators over the repeats give $20.0\%$ and $18.8\%$, both also significant.\footnote{Four of the five dimensions individually run in this direction; Constrained Generation is noisier under repetition and runs against it, and excluding it shrinks the comparison to $15.0\%$ vs.\ $14.4\%$.} Spending a fixed evaluation budget on diagnostic breadth therefore surfaces more failures than spending it on depth, which is the central practical argument for evaluating several uses of the same inferred rule.

\subsection{Effects of Generative Sampling}

A key advantage of PotARCin is that generator programs allow us to sample many distinct input/output pairs for the same underlying ARC task, conditioned on the same demonstrations. This lets us test whether apparent skill competence is robust across multiple instantiations, rather than only across repeated model calls with different stochastic outputs. In the main evaluation, each task is evaluated using a single sampled configuration for each dimension. Here, we ask whether tasks that appear solved under this limited sampling budget remain solved when evaluated more extensively.

Because this experiment is more resource-intensive, we restrict it to GPT-5.4 on a small subset of ARC-AGI-1 training tasks. We begin with tasks for which GPT-5.4 passes all five PotARCin dimensions under the main evaluation. 
From these, we select 50 random tasks and filter out cases where the generated Definition program implements only trivially simple operations, such as a single rotation, duplication, or basic color remapping, leaving 20 tasks for the increased-sampling study.

For each of these 20 tasks, we sample 10 distinct multi-dimensional evaluation configurations. Classification, Editing, and Inversion naturally support this procedure because each can be resampled by drawing new candidate pairs, corruptions, or target outputs. Definition and Constrained Generation require slight modifications, since repeated calls do not necessarily produce meaningfully different evaluation instances. For Definition, we do not ask the model to write a new program; instead, we increase the number of dynamic generator samples used to test the original Definition program, from 50 to 1,000. For Constrained Generation, we test whether the model can produce multiple distinct valid examples. Whenever the model generates a valid pair, we add it to the demonstrations for the next generation attempt and instruct the model not to copy or minimally alter previous examples.

\begin{figure}[htbp]
  \centering
  \includegraphics[width=0.6\textwidth]{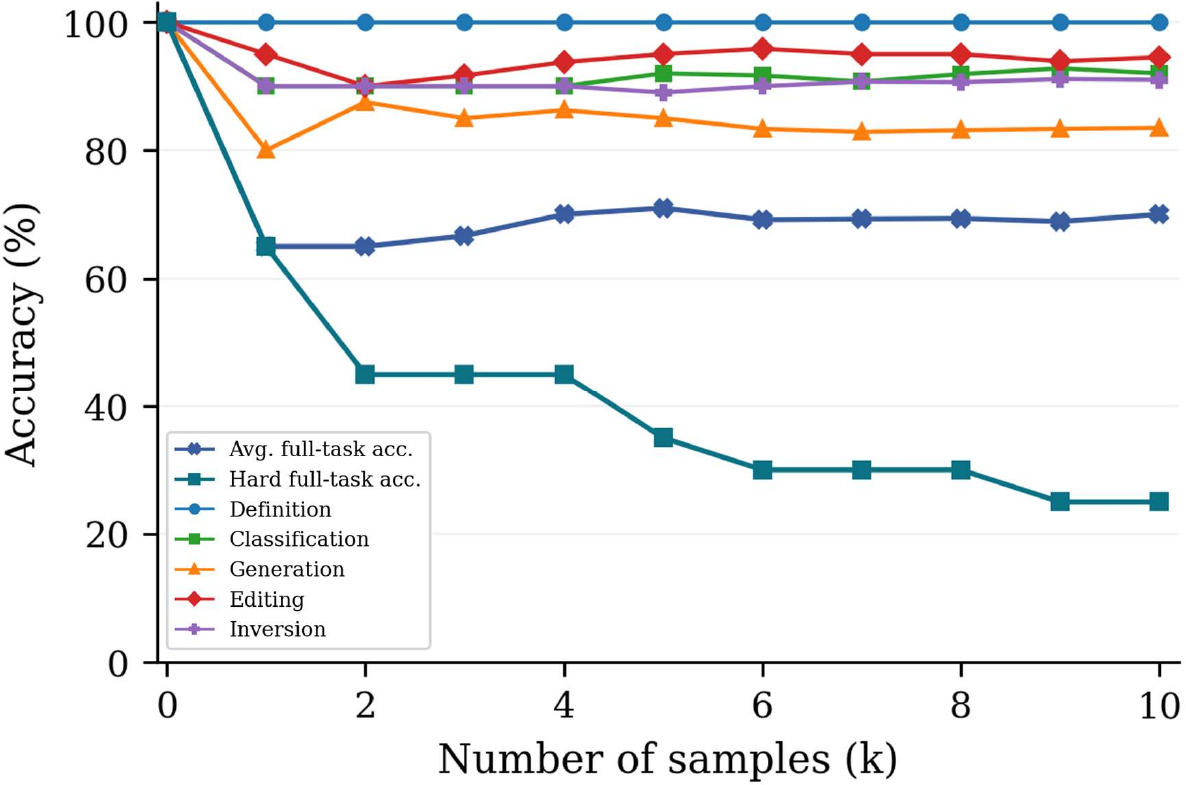}
  \caption{Results of an increased sampling budget across 20 ARC-AGI-1 training tasks that GPT-5.4 initially solved across all five PotARCin dimensions. ``Hard full-task accuracy'' is the percentage of tasks passing \emph{all} sampled configurations up to that budget.}
  \label{fig:sampling_overview}
\end{figure}

Results are shown in \autoref{fig:sampling_overview}, with statistics for selected sampling budgets reported in \autoref{tab:gpt54_subset20_sampling_budget_trends}. On 15 of the 20 tasks, GPT-5.4 fails at least one of the 10 sampled configurations, despite having passed all five dimensions in the main evaluation. Averaged across the selected tasks, full-task accuracy under increased sampling is 70\%. These results suggest that single-sample full-task accuracy can still overestimate robust rule competence: a model may pass all five dimensions once, yet fail when the same rule is probed through additional valid instantiations. 
For Definition, increasing the number of dynamic test samples produces little additional change: even with up to 10,000 additional samples, the initial pass/fail outcome never changes. For the other resampled dimensions, failures tend to emerge quickly, with 11 of the 15 affected tasks exhibiting at least one failure within the first two additional samples. One caveat is that this experiment uses GPT-5.4 with reasoning enabled, for which the OpenAI API does not expose the temperature parameter. Sampling stochasticity is therefore present, but cannot be separated from variation due to resampling the task instance.

\subsection{Difficulty of Corruption Types}

To better understand the Classification dimension, we compute model failure rates separately for each corruption type on both evaluation datasets. Results are shown in \autoref{tab:failure_rates_by_corruption}. Each entry is the fraction of examples of a given type that the model mislabels.

Human-generated errors are, in general, the most difficult corrupted examples for models to classify. This is especially pronounced on P-ARC, where failure rates on human corruptions are substantially higher than for other corruption types. This pattern is intuitive: unlike synthetic perturbations such as color flips or mismatched instances, human errors often arise from attempts to apply a related but incorrect interpretation of the task rule, or from subtle mistakes that are difficult to detect. We also observe substantial differences between models, with GPT-5.4 achieving the lowest failure rate in 8 of the 12 corruption-type/dataset categories.

Across datasets, the relative difficulty of corruption types is broadly similar, but average failure rates are higher on P-ARC. This likely reflects both the greater difficulty of the held-out tasks and differences in corruption quality. In particular, P-ARC human corruptions were created specifically for our tasks, whereas H-ARC likely includes many human outputs that are clearly incorrect. The retrieval corruption type may also be affected by distribution shift: because the retrieval database is built from ARC-AGI-1 tasks, retrieved examples may be less closely matched to P-ARC tasks than to public ARC-AGI-1 tasks.


Because Classification is scored strictly, it admits two trivial baselines worth stating. The candidate mixture is roughly $71\%$ invalid, so always answering ``invalid'' scores $19.1\%$ on ARC-AGI-1 and $16.0\%$ on P-ARC, while always answering ``valid'' scores $0.3\%$ and $2.0\%$. Every model clears the always-invalid baseline comfortably on ARC-AGI-1. On P-ARC it is closer: Gemini 3.1 Pro ($19.0\%$) is barely above it and MiniMax M2.5 ($14.0\%$) falls below, so P-ARC Classification accuracy should not be read as evidence of rule-consistent discrimination for the weaker models.

\begin{table}[htbp]
\centering
\caption{\textbf{Failure rates by corruption type and dataset}, pooled over both runs. Entries are percentages: the fraction of candidates of each type the model labelled incorrectly. For corrupted candidates, failure means accepting an invalid pair; for valid candidates (``Correct'') it means rejecting a valid pair. Bold marks the lowest rate in each column. The last row gives the median number of candidates of that type per model; counts vary by $\le 1\%$ across models. Wilson $95\%$ intervals are in \autoref{tab:failure_rates_ci}.}
\label{tab:failure_rates_by_corruption}
\resizebox{\columnwidth}{!}{
\begin{tabular}{lcccccccccccc}
\toprule
        & \multicolumn{6}{c}{\textbf{ARC-AGI-1}}
        & \multicolumn{6}{c}{\textbf{P-ARC}} \\
        \cmidrule(lr){2-7} \cmidrule(lr){8-13}
        \textbf{Model}
        & \makecell{\textbf{Human}} & \makecell{\textbf{Corrupted}\\ \textbf{verifier}} & \textbf{Correct}
        & \makecell{\textbf{Color}\\ \textbf{flip}} & \textbf{Retrieval} & \makecell{\textbf{Instance}\\ \textbf{mismatch}}
        & \makecell{\textbf{Human}} & \makecell{\textbf{Corrupted}\\ \textbf{verifier}} & \textbf{Correct}
        & \makecell{\textbf{Color}\\ \textbf{flip}} & \textbf{Retrieval} & \makecell{\textbf{Instance}\\ \textbf{mismatch}} \\
\midrule
Claude 4.6 Opus & 20.74 & 11.14 & \textbf{1.84} & 8.76 & 4.52 & 4.59 & 65.03 & 12.50 & \textbf{4.93} & 28.57 & 7.14 & 12.22 \\
GPT-5.4 & \textbf{8.21} & \textbf{2.30} & 5.06 & 6.53 & 3.15 & \textbf{1.85} & \textbf{47.22} & \textbf{0.00} & 16.67 & \textbf{0.00} & \textbf{0.00} & \textbf{3.26} \\
Gemini 3.1 Pro & 16.37 & 8.99 & 5.48 & \textbf{2.82} & \textbf{2.17} & 6.81 & 60.42 & 9.38 & 22.92 & 14.29 & 2.38 & 7.61 \\
Kimi K2.5 & 18.54 & 11.24 & 5.98 & 25.78 & 10.79 & 7.75 & 61.11 & 14.06 & 12.59 & 42.86 & 14.29 & 11.83 \\
MiniMax M2.5 & 31.32 & 26.62 & 3.66 & 40.23 & 40.43 & 18.12 & 70.83 & 23.44 & 12.50 & 50.00 & 30.95 & 23.91 \\
\midrule
\textit{items per model} & 562 & 432 & 1146 & 353 & 417 & 1086 & 144 & 64 & 144 & 14 & 42 & 92 \\
\bottomrule
\end{tabular}}
\end{table}

\subsection{Self-Consistency Conditioned on a Correct Definition}\label{sec:selfcons}

\citet{pmlr-v267-mancoridis25a} evaluate self-consistency by comparing model responses across related probes of the same concept and identifying cases where those responses disagree. PotARCin enables an analogous test for ARC. Because the Definition dimension asks models to produce an executable program, we can compare that program against the model's responses in other dimensions: for Constrained Generation and Inversion, we check whether the Definition program maps the model-proposed input grid to its output grid; for Classification, we compare the model's valid/invalid judgment with whether the Definition program maps the candidate input to its output.

To distinguish genuine cross-dimensional inconsistency from cases in which the model's Definition is itself wrong, we additionally condition this analysis on a correct executable Definition. Here, a correct Definition is a loadable program that passes the task's training and test examples. We then compare that program with Classification, Constrained Generation, and Inversion responses, pooling both runs for all five models. A full overview is shown in \autoref{tab:selfcons_conditioned}.

When both the Definition and the response in the other dimension are correct, pooled agreement is 99.1\% on ARC-AGI-1 and 94.0\% on P-ARC. In contrast, when the Definition is correct but the other response is incorrect, agreement falls to 21.1\% on ARC-AGI-1 and 17.1\% on P-ARC. Thus, most mistakes in the other dimensions cannot be explained by the model consistently applying its own correctly formalized rule. The effect is strongest for Classification and Constrained Generation, while Inversion varies more across models; the P-ARC estimates should be interpreted cautiously because the conditional denominators are smaller. Unconditional agreement rates, without conditioning on Definition correctness, are reported in \autoref{tab:self_consistency_combined}.

These conditioned inconsistencies are direct evidence of a Potemkin-like failure mode: models frequently contradict a rule they have correctly formalized in another dimension. Unlike the pre-existing concepts studied by \citet{pmlr-v267-mancoridis25a}, however, the object of understanding here is a task-specific skill: a transformation rule inferred from a few demonstrations.

\section{Conclusion}

We introduced PotARCin, a benchmark that extends ARC evaluation beyond output-grid correctness by probing five dimensions of rule competence: Definition, Classification, Constrained Generation, Editing, and Inversion. Full-task accuracy is 25--52 percentage points below output-grid accuracy on ARC-AGI-1 and only 1--8\% on P-ARC. Under increased generative sampling, GPT-5.4 fails at least one of ten resampled configurations on 15 of 20 tasks it initially passed, and human-generated errors are the hardest corruptions to classify. Probing five dimensions once exposes more failures than sampling one dimension five times (25.5\% vs.\ 20.2\%). Given a correct executable Definition, agreement is 99.1\% (ARC-AGI-1) and 94.0\% (P-ARC) for correct responses in other dimensions, but only 21.1\% and 17.1\% for incorrect ones, pointing to a Potemkin-like failure mode: models may correctly formalize a rule yet fail to apply it consistently.

Overall, our results suggest that standard ARC accuracy provides an incomplete picture of abstract skill acquisition. Producing the correct output grid for a single test input does not necessarily imply that a model has acquired a reusable, flexible representation of the underlying transformation rule. Multi-dimensional evaluation also reorders models that standard scores rank alike, and the weakest model on output grids is not the weakest on broader rule competence. More generally, PotARCin supports evaluating not only depth within a fixed task format, but also breadth across different uses of the same inferred rule. Even on ARC-AGI-1 training tasks, the observed drops in full-task accuracy and self-consistency suggest that ARC-AGI-1 should not be regarded as fully ``solved'' from the perspective of abstract skill acquisition. While newer benchmarks such as ARC-AGI-3~\citep{foundation2026arcagi3newchallengefrontier} explore richer interactive settings, our results show that even simple static transformation rules remain challenging when models are asked to use them flexibly. By combining generator- and verifier-based evaluation with multiple rule-use dimensions, PotARCin offers a more demanding and more diagnostic framework for measuring abstract rule competence in ARC-style tasks.

\textbf{Limitations and Future Work.} PotARCin depends on the quality of its generator and verifier programs. Although these programs enable scalable automated evaluation, they must remain aligned with each task's intended transformation rule. For P-ARC, we write and test these programs; for ARC-AGI-1, we reuse existing verifier sources and observe that some candidate verifiers capture nearby variants rather than the intended rule. Even with extensive testing, exact rule alignment is difficult to guarantee. A related, more fundamental issue concerns Classification: ARC tasks are underspecified and may admit multiple rules consistent with the demonstrations~\citep{beger2026aimodelsperformhumanlike}. A pair labeled invalid under the designated verifier may therefore remain explicable under an alternative consistent rule. Classification should be interpreted as testing adherence to the benchmark's intended rule, not as proving that no consistent alternative exists. Another limitation is model-compute comparability. Although all models are configured for low reasoning effort, Claude Opus 4.6 was observed to consume substantially more tokens than other models on P-ARC, which may affect comparisons; see also \autoref{sec:adaptive_token}. In addition, P-ARC's estimated difficulty rests on an internal feasibility check rather than a formal human study; formal solve-rate and solve-time measurements are left to future work. Finally, several aspects of rule competence remain outside our scope. We restrict generated examples to the mimetic similarity boundary of the demonstrations, leaving principled rule shifts and conceptual slippage~\citep{hofstadter2019conceptual} for future work. We also observe that models often exploit the Constrained Generation and Inversion dimensions by producing the simplest valid examples possible. Future work could quantify the diversity of model-generated examples, especially for autonomous ARC-style task construction.

\begin{ack}
This research was supported by the BANYAN Institute, funded by Sandia National Laboratories’ Laboratory Directed Research and Development program, and by the grant "Building Diverse Intelligences through Compositionality and Mechanism Design" from the Templeton World Charity Foundation, https://doi.org/10.54224/33804 and https://doi.org/10.54224/20650.  The authors would like to thank Marina Mancoridis for visiting the Santa Fe Institute and presenting her work on Potemkin Understanding, and Brenden Lake and Thomas Griffiths for constructive discussions regarding the PotARCin experiments. We also thank Anna Patelli for supporting the development of the P-ARC test set and Navya Sahay for assistance with model evaluation.

Sandia National Laboratories is a multimission laboratory managed and operated by National Technology and Engineering Solutions of Sandia, LLC., a wholly owned subsidiary of Honeywell International, Inc., for the U.S. Department of Energy’s National Nuclear Security Administration under contract DE-NA-0003525. The authors declare no competing interests.
\end{ack}

\newpage
\bibliographystyle{plainnat}
\bibliography{bibliography}

@InProceedings{pmlr-v267-mancoridis25a,
  title = 	 {Potemkin Understanding in Large Language Models},
  author =       {Mancoridis, Marina and Weeks, Bec and Vafa, Keyon and Mullainathan, Sendhil},
  booktitle = 	 {Proceedings of the 42nd International Conference on Machine Learning},
  pages = 	 {42857--42881},
  year = 	 {2025},
  editor = 	 {Singh, Aarti and Fazel, Maryam and Hsu, Daniel and Lacoste-Julien, Simon and Berkenkamp, Felix and Maharaj, Tegan and Wagstaff, Kiri and Zhu, Jerry},
  volume = 	 {267},
  series = 	 {Proceedings of Machine Learning Research},
  month = 	 {13--19 Jul},
  publisher =    {PMLR},
  url = 	 {https://proceedings.mlr.press/v267/mancoridis25a.html}
}

@misc{chollet2019measureintelligence,
      title={On the Measure of Intelligence}, 
      author={François Chollet},
      year={2019},
      eprint={1911.01547},
      archivePrefix={arXiv},
      primaryClass={cs.AI},
      url={https://arxiv.org/abs/1911.01547}, 
}

@misc{chollet2026arcagi2newchallengefrontier,
      title={ARC-AGI-2: A New Challenge for Frontier AI Reasoning Systems}, 
      author={Francois Chollet and Mike Knoop and Gregory Kamradt and Bryan Landers and Henry Pinkard},
      year={2026},
      eprint={2505.11831},
      archivePrefix={arXiv},
      primaryClass={cs.AI},
      url={https://arxiv.org/abs/2505.11831}, 
}

@misc{foundation2026arcagi3newchallengefrontier,
      title={ARC-AGI-3: A New Challenge for Frontier Agentic Intelligence}, 
      author={ARC Prize Foundation},
      year={2026},
      eprint={2603.24621},
      archivePrefix={arXiv},
      primaryClass={cs.AI},
      url={https://arxiv.org/abs/2603.24621}, 
}

@misc{chollet2025arcprize2024technical,
      title={ARC Prize 2024: Technical Report}, 
      author={Francois Chollet and Mike Knoop and Gregory Kamradt and Bryan Landers},
      year={2025},
      eprint={2412.04604},
      archivePrefix={arXiv},
      primaryClass={cs.AI},
      url={https://arxiv.org/abs/2412.04604}, 
}

@misc{beger2026aimodelsperformhumanlike,
      title={Do AI Models Perform Human-like Abstract Reasoning Across Modalities?}, 
      author={Claas Beger and Ryan Yi and Shuhao Fu and Kaleda Denton and Arseny Moskvichev and Sarah W. Tsai and Sivasankaran Rajamanickam and Melanie Mitchell},
      year={2026},
      eprint={2510.02125},
      archivePrefix={arXiv},
      primaryClass={cs.AI},
      url={https://arxiv.org/abs/2510.02125}, 
}

@misc{mineault2026cognitivedarkmattermeasuring,
      title={Cognitive Dark Matter: Measuring What AI Misses}, 
      author={Patrick J. Mineault and Thomas L. Griffiths and Sean Escola},
      year={2026},
      eprint={2603.03414},
      archivePrefix={arXiv},
      primaryClass={q-bio.NC},
      url={https://arxiv.org/abs/2603.03414}, 
}

@inproceedings{ribeiro-etal-2020-beyond,
    title = "Beyond Accuracy: Behavioral Testing of {NLP} Models with {C}heck{L}ist",
    author = "Ribeiro, Marco Tulio  and
      Wu, Tongshuang  and
      Guestrin, Carlos  and
      Singh, Sameer",
    editor = "Jurafsky, Dan  and
      Chai, Joyce  and
      Schluter, Natalie  and
      Tetreault, Joel",
    booktitle = "Proceedings of the 58th Annual Meeting of the Association for Computational Linguistics",
    month = jul,
    year = "2020",
    address = "Online",
    publisher = "Association for Computational Linguistics",
    url = "https://aclanthology.org/2020.acl-main.442/",
    doi = "10.18653/v1/2020.acl-main.442",
    pages = "4902--4912"
}

@article{li2024combining,
  title={Combining induction and transduction for abstract reasoning},
  author={Li, Wen-Ding and Hu, Keya and Larsen, Carter and Wu, Yuqing and Alford, Simon and Woo, Caleb and Dunn, Spencer M and Tang, Hao and Naim, Michelangelo and Nguyen, Dat and others},
  journal={arXiv preprint arXiv:2411.02272},
  year={2024}
}

@misc{pourcel2026selfimprovinglanguagemodelsevolutionary,
      title={Self-Improving Language Models for Evolutionary Program Synthesis: A Case Study on ARC-AGI}, 
      author={Julien Pourcel and Cédric Colas and Pierre-Yves Oudeyer},
      year={2026},
      eprint={2507.14172},
      archivePrefix={arXiv},
      primaryClass={cs.LG},
      url={https://arxiv.org/abs/2507.14172}, 
}

@misc{hodel2024addressingabstractionreasoningcorpus,
      title={Addressing the Abstraction and Reasoning Corpus via Procedural Example Generation}, 
      author={Michael Hodel},
      year={2024},
      eprint={2404.07353},
      archivePrefix={arXiv},
      primaryClass={cs.LG},
      url={https://arxiv.org/abs/2404.07353}, 
}

@misc{moffitt2025arcgenmimeticproceduralbenchmark,
      title={ARC-GEN: A Mimetic Procedural Benchmark Generator for the Abstraction and Reasoning Corpus}, 
      author={Michael D. Moffitt},
      year={2025},
      eprint={2511.00162},
      archivePrefix={arXiv},
      primaryClass={cs.AI},
      url={https://arxiv.org/abs/2511.00162}, 
}

@misc{google-code-golf-2025,
    author = {Michael D. Moffitt and Divy Thakkar and Ryan Burnell and Orhan Firat and Walter Reade and Sohier Dane and Addison Howard},
    title = {NeurIPS 2025 - Google Code Golf Championship},
    year = {2025},
    howpublished = {\url{https://kaggle.com/competitions/google-code-golf-2025}},
    note = {Kaggle}
}

@misc{hu2025arcvisionproblem,
      title={ARC Is a Vision Problem!}, 
      author={Keya Hu and Ali Cy and Linlu Qiu and Xiaoman Delores Ding and Runqian Wang and Yeyin Eva Zhu and Jacob Andreas and Kaiming He},
      year={2025},
      eprint={2511.14761},
      archivePrefix={arXiv},
      primaryClass={cs.CV},
      url={https://arxiv.org/abs/2511.14761}, 
}

@article{LeGris2025,
  title = {A Comprehensive Behavioral Dataset for the Abstraction and Reasoning Corpus},
  volume = {12},
  ISSN = {2052-4463},
  url = {http://dx.doi.org/10.1038/s41597-025-05687-1},
  DOI = {10.1038/s41597-025-05687-1},
  number = {1},
  journal = {Scientific Data},
  publisher = {Springer Science and Business Media LLC},
  author = {LeGris,  Solim and Vong,  Wai Keen and Lake,  Brenden M. and Gureckis,  Todd M.},
  year = {2025},
  month = Aug 
}

@misc{arcprizeOpenAIBreakthrough,
	author = {ARC Prize Foundation},
	title = {{O}pen{A}{I} o3 {B}reakthrough {H}igh {S}core on {A}{R}{C}-{A}{G}{I}-{P}ub | {A}{R}{C} {P}rize --- arcprize.org},
	howpublished = {\url{https://arcprize.org/blog/oai-o3-pub-breakthrough}},
	year = {2025},
	note = {[Accessed 06-05-2026]},
}

@misc{Anthropic, title={System card: Claude Opus 4.6 February 2026 anthropic.com}, url={https://www-cdn.anthropic.com/0dd865075ad3132672ee0ab40b05a53f14cf5288.pdf}, journal={Claude Opus 4.6 System Card}, publisher={Anthropic}, author={Anthropic}}

@misc{srivastava2023imitationgamequantifyingextrapolating,
      title={Beyond the Imitation Game: Quantifying and extrapolating the capabilities of language models}, 
      author={Srivastava, Aarohi and Rastogi, Abhinav and Rao, Abhishek and Shoeb, Abu Awal Md and Abid, Abubakar and Fisch, Adam and Brown, Adam R. and Santoro, Adam and Gupta, Aditya and Garriga-Alonso, Adri{\`a} and others},
      year={2023},
      eprint={2206.04615},
      archivePrefix={arXiv},
      primaryClass={cs.CL},
      url={https://arxiv.org/abs/2206.04615}, 
}

@misc{mirzadeh2025gsmsymbolicunderstandinglimitationsmathematical,
      title={GSM-Symbolic: Understanding the Limitations of Mathematical Reasoning in Large Language Models}, 
      author={Iman Mirzadeh and Keivan Alizadeh and Hooman Shahrokhi and Oncel Tuzel and Samy Bengio and Mehrdad Farajtabar},
      year={2025},
      eprint={2410.05229},
      archivePrefix={arXiv},
      primaryClass={cs.LG},
      url={https://arxiv.org/abs/2410.05229}, 
}

@misc{sonwane2026omnicodebenchmarkevaluatingsoftware,
      title={OmniCode: A Benchmark for Evaluating Software Engineering Agents}, 
      author={Atharv Sonwane and Eng-Shen Tu and Wei-Chung Lu and Claas Beger and Carter Larsen and Debjit Dhar and Simon Alford and Rachel Chen and Ronit Pattanayak and Tuan Anh Dang and Guohao Chen and Gloria Geng and Kevin Ellis and Saikat Dutta},
      year={2026},
      eprint={2602.02262},
      archivePrefix={arXiv},
      primaryClass={cs.SE},
      url={https://arxiv.org/abs/2602.02262}, 
}

@inproceedings{hofstadter2019conceptual,
  title={Conceptual slippage and analogy-making: a report on the copycat project},
  author={Hofstadter, Douglas R and Mitchell, Melanie},
  booktitle={10th Annual Conference Cognitive Science Society Pod},
  pages={601--607},
  year={2019},
  organization={Psychology Press}
}

@misc{ARC-AGI-1,
author = {Chollet, François},
title = {{The Abstraction and Reasoning Corpus (ARC)}},
year ={2026},
note = {\url{https://github.com/fchollet/ARC}, last accessed, February 2026}
}

@article{moskvichev2023conceptarc,
  title={The {ConceptARC} benchmark: Evaluating understanding and generalization in the {ARC} domain},
  author={Moskvichev, Arseny and Odouard, Victor Vikram and Mitchell, Melanie},
  journal={Transactions on Machine Learning Research},
  year={2023}
}

@misc{deliège2026implicitruleinductiontesttime,
      title={Implicit Rule Induction with Test-Time Task Embeddings in ARC-like Tasks}, 
      author={Adrien Deliège and Claas Beger and Marc Van Droogenbroeck and Melanie Mitchell},
      year={2026},
      eprint={2609.21181},
      archivePrefix={arXiv},
      primaryClass={cs.AI},
      url={https://arxiv.org/abs/2609.21181}, 
}

\newpage
\appendix


\section{Corruption Type Examples}
\label{sec:corruption_types}
\begin{figure}[htbp]
  \centering

  \begin{minipage}{0.32\textwidth}
    \centering
    \includegraphics[width=\linewidth]{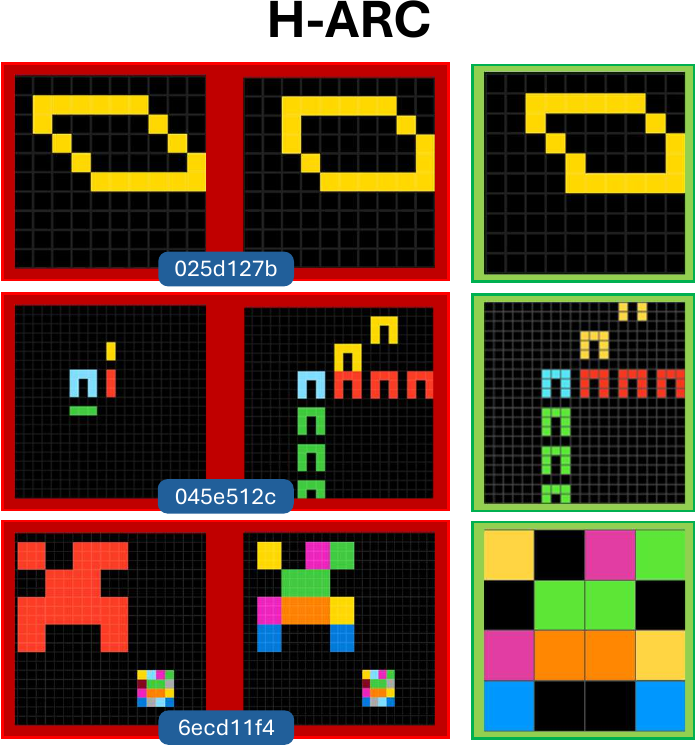}
  \end{minipage}
  \hfill
  \begin{minipage}{0.32\textwidth}
    \centering
    \includegraphics[width=\linewidth]{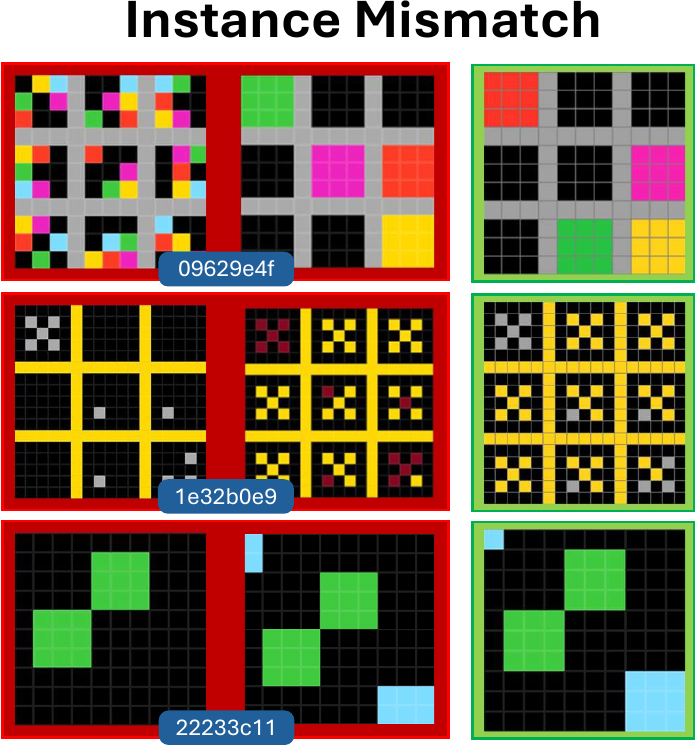}
  \end{minipage}
  \hfill
  \begin{minipage}{0.32\textwidth}
    \centering
    \includegraphics[width=\linewidth]{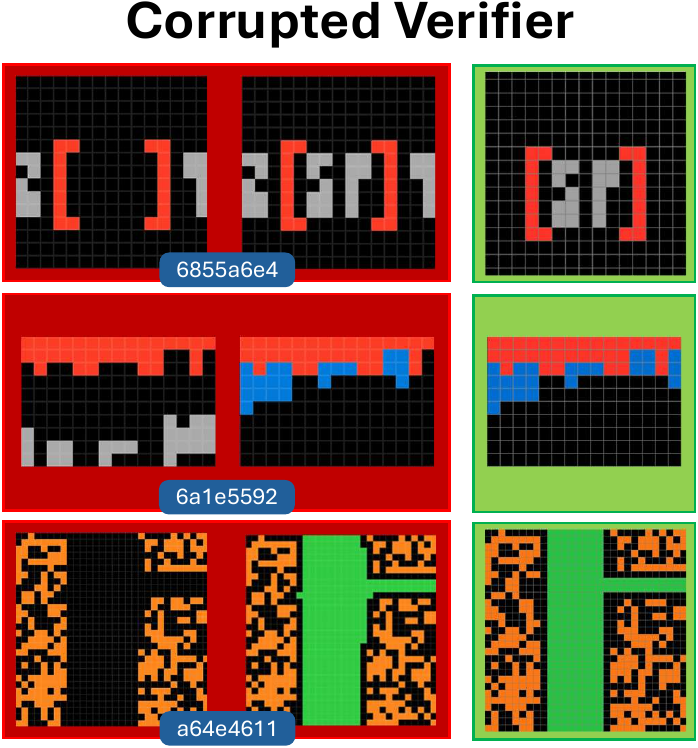}
  \end{minipage}

  \caption{Examples of corruption types generated on the ARC-AGI-1 training set. Correct output grids are shown to the right of each corrupted pair, outlined in green. \textbf{Left:} H-ARC corruptions are errors made by human participants in the H-ARC dataset. \textbf{Middle:} Instance mismatch corruptions replace the correct output grid with a similar output from the generator dataset, selected using an edit-distance metric. \textbf{Right:} Corrupted verifiers produce outputs from perturbed verifier programs, either by deleting AST branches or reassigning intermediate variables.}
  \label{fig:Corruption_Sources}
\end{figure}

\begin{figure}[htbp]
  \centering

  \includegraphics[width=0.7\textwidth]{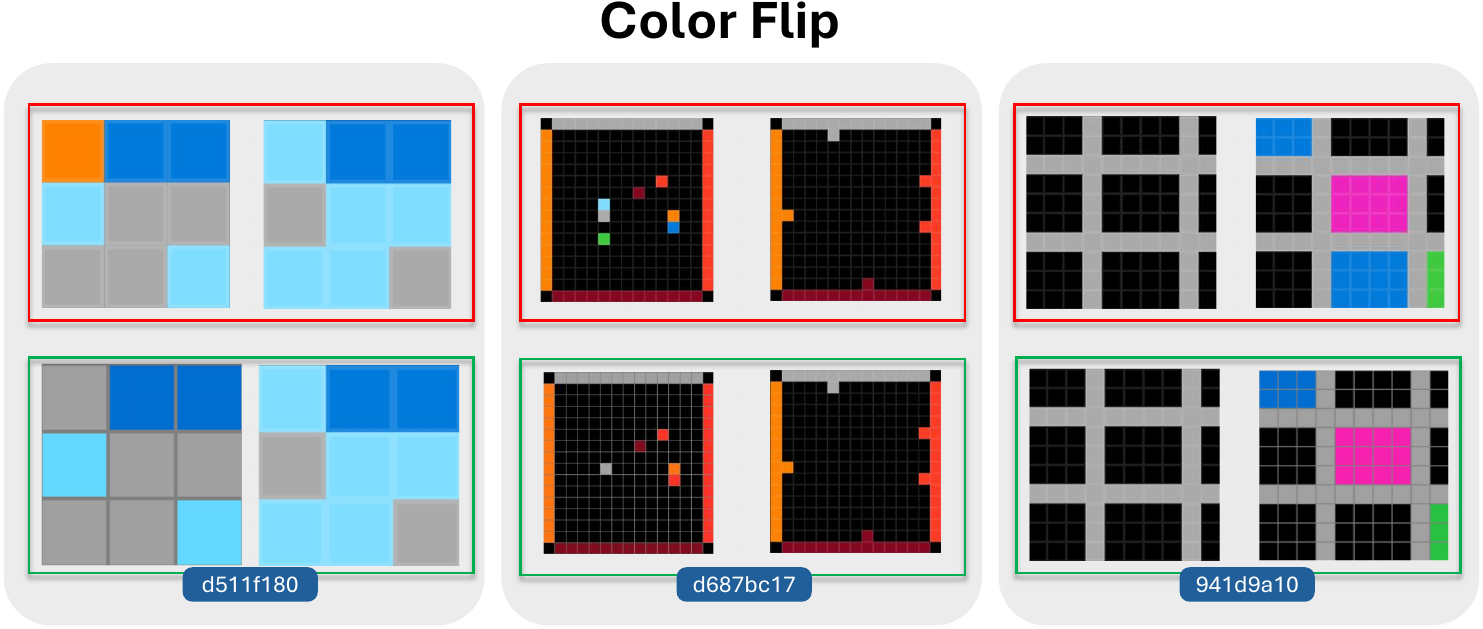}

  \vspace{0.5em}

  \includegraphics[width=0.7\textwidth]{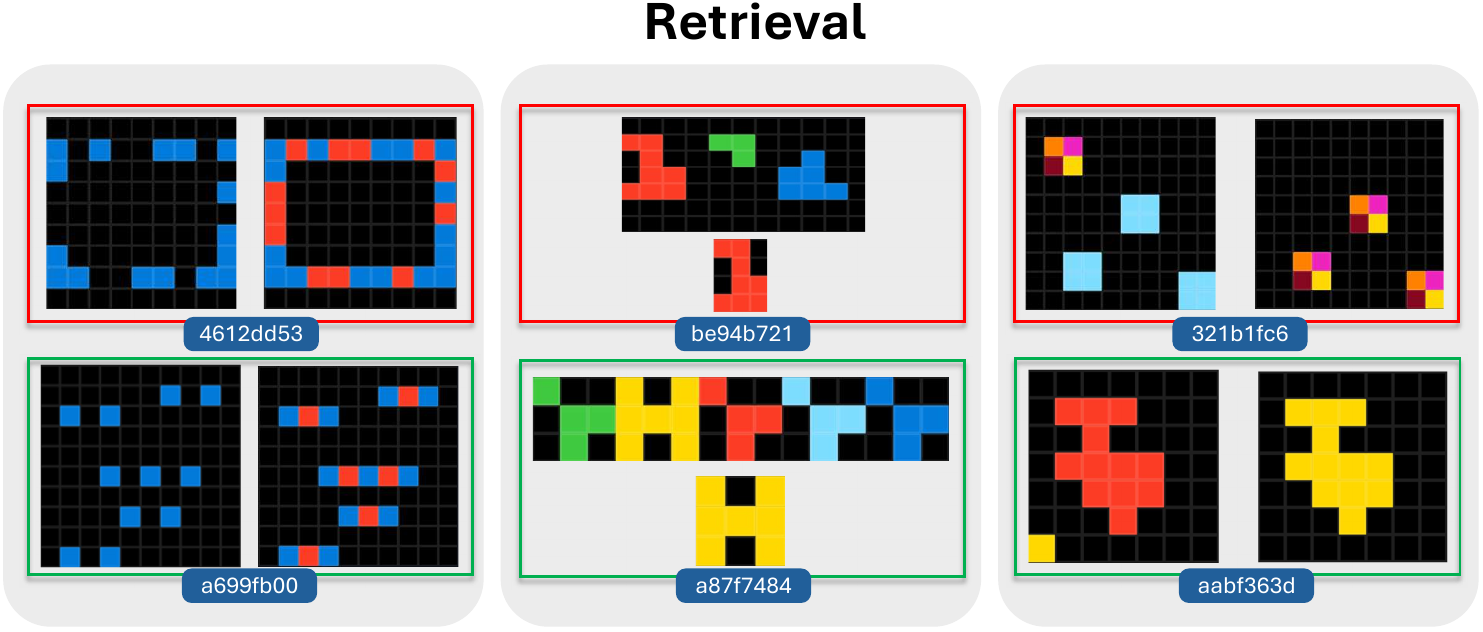}

  \caption{Examples of color flip and retrieval corruptions. \textbf{Top:} Color flip corruptions change the color of a sampled cell or small connected group in either the input or output grid. \textbf{Bottom:} Retrieval corruptions use embedding-based lookup to retrieve tasks with similar underlying rules, from which an input-output pair is sampled.}
  \label{fig:overview_vertical}
\end{figure}

\clearpage

\section{Sampling Mixture for Classification and Editing}

\begin{table}[htbp]
\centering
\caption{Sampling weights over valid pairs and corruption types used to construct Classification candidates. The weights differ between datasets to reflect differences in corruption quality: Retrieval is less suitable for the smaller, distribution-shifted P-ARC set (lower embedding-similarity scores), and Color Flip is less challenging because P-ARC contains fewer color-transformation rules, whereas manually validated P-ARC verifiers and human errors are comparatively reliable. The overall split between valid and invalid candidates is held fixed.}
\label{tab:corruption_weights}
\footnotesize
\begin{tabular}{lcc}
\toprule
\textbf{Candidate type} & \textbf{ARC-AGI-1} & \textbf{P-ARC} \\
\midrule
Valid (dynamic correct)      & 0.250 & 0.250 \\
Human error (H-ARC / P-ARC)  & 0.235 & 0.265 \\
Instance mismatch            & 0.215 & 0.215 \\
Corrupted verifier           & 0.130 & 0.160 \\
Retrieval                    & 0.103 & 0.073 \\
Color flip                   & 0.067 & 0.037 \\
\bottomrule
\end{tabular}
\end{table}

\clearpage

\section{Additional Experimental Results}

\begin{table*}[htbp]
\centering
\caption{Model performance on ARC-AGI-1 Train and P-ARC, pooled over two independent runs ($n{=}800$ and $n{=}100$ task evaluations). Entries are percentages; brackets are pooled Wilson $95\%$ intervals. \textbf{Indep.} is the product of the five dimension accuracies, i.e.\ the full-task rate expected if per-dimension pass events were independent. Bold marks the best full-task mean within each dataset block. These are the exact values underlying \autoref{fig:performance_overview}.}
\label{tab:main_results}
\scriptsize
\setlength{\tabcolsep}{3.0pt}
\renewcommand{\arraystretch}{1.15}
\resizebox{\textwidth}{!}{%
\begin{tabular}{lcccccccc}
\toprule
\textbf{Model} & \makecell{\textbf{Output grid}} & \textbf{Definition} & \textbf{Classification} & \textbf{Generation} & \textbf{Editing} & \textbf{Inversion} & \textbf{Full-task} & \textbf{Indep.} \\
\midrule
\multicolumn{9}{l}{\textsc{ARC-AGI-1 Train}}\\
GPT-5.4 Low & 84.2 \ci{81.6}{86.6} & 73.6 \ci{70.5}{76.6} & 83.8 \ci{81.0}{86.1} & 77.2 \ci{74.2}{80.0} & 84.0 \ci{81.3}{86.4} & 85.1 \ci{82.5}{87.4} & 46.6 \ci{43.2}{50.1} & 34.0 \\
Gemini 3.1 Pro Low & 89.1 \ci{86.8}{91.1} & 79.2 \ci{76.3}{81.9} & 75.9 \ci{72.8}{78.7} & 88.0 \ci{85.6}{90.1} & 85.8 \ci{83.2}{88.0} & 88.2 \ci{85.8}{90.3} & \textbf{58.2} \ci{54.8}{61.6} & 40.0 \\
Claude 4.6 Opus Low (120K) & 87.6 \ci{85.2}{89.7} & 66.8 \ci{63.4}{69.9} & 71.5 \ci{68.3}{74.5} & 78.9 \ci{75.9}{81.6} & 88.2 \ci{85.8}{90.3} & 91.0 \ci{88.8}{92.8} & 37.6 \ci{34.3}{41.0} & 30.2 \\
Kimi K2.5 & 63.4 \ci{60.0}{66.6} & 64.5 \ci{61.1}{67.7} & 63.5 \ci{60.1}{66.8} & 71.8 \ci{68.5}{74.8} & 79.6 \ci{76.7}{82.3} & 85.8 \ci{83.2}{88.0} & 38.6 \ci{35.3}{42.0} & 20.1 \\
MiniMax M2.5 & 72.9 \ci{69.7}{75.8} & 56.6 \ci{53.2}{60.0} & 42.4 \ci{39.0}{45.8} & 64.6 \ci{61.2}{67.9} & 68.6 \ci{65.3}{71.7} & 76.6 \ci{73.6}{79.4} & 20.6 \ci{18.0}{23.6} & 8.1 \\
\midrule
\multicolumn{9}{l}{\textsc{P-ARC}}\\
GPT-5.4 Low & 31.0 \ci{22.8}{40.6} & 14.0 \ci{8.5}{22.1} & 36.0 \ci{27.3}{45.8} & 39.0 \ci{30.0}{48.8} & 34.0 \ci{25.5}{43.7} & 47.0 \ci{37.5}{56.7} & 6.0 \ci{2.8}{12.5} & 0.3 \\
Gemini 3.1 Pro Low & 45.0 \ci{35.6}{54.8} & 12.0 \ci{7.0}{19.8} & 19.0 \ci{12.5}{27.8} & 42.0 \ci{32.8}{51.8} & 32.0 \ci{23.7}{41.7} & 50.0 \ci{40.4}{59.6} & 4.0 \ci{1.6}{9.8} & 0.2 \\
Claude 4.6 Opus Low (120K) & 64.0 \ci{54.2}{72.7} & 25.0 \ci{17.5}{34.3} & 27.0 \ci{19.3}{36.4} & 56.0 \ci{46.2}{65.3} & 58.0 \ci{48.2}{67.2} & 71.0 \ci{61.5}{79.0} & \textbf{8.0} \ci{4.1}{15.0} & 1.6 \\
Kimi K2.5 & 17.0 \ci{10.9}{25.5} & 10.0 \ci{5.5}{17.4} & 25.0 \ci{17.5}{34.3} & 39.0 \ci{30.0}{48.8} & 37.0 \ci{28.2}{46.8} & 47.0 \ci{37.5}{56.7} & 2.0 \ci{0.6}{7.0} & 0.2 \\
MiniMax M2.5 & 24.0 \ci{16.7}{33.2} & 6.0 \ci{2.8}{12.5} & 14.0 \ci{8.5}{22.1} & 23.0 \ci{15.8}{32.2} & 17.0 \ci{10.9}{25.5} & 29.0 \ci{21.0}{38.5} & 1.0 \ci{0.2}{5.4} & 0.0 \\
\bottomrule
\end{tabular}}
\end{table*}

\FloatBarrier

\subsection{Constrained Generation Excluding Stable-Set Matches}\label{sec:cg_strict}

Constrained Generation rejects copies of the demonstrations shown to the model, but not exact matches to examples in the ARC-GEN stable set. As a stricter control, we exclude such matches. This lowers ARC-AGI-1 Constrained Generation accuracy to 70.0\% for GPT-5.4 ($-7.2$), 80.4\% for Gemini 3.1 Pro ($-7.6$), 73.5\% for Claude Opus 4.6 ($-5.4$), 67.6\% for Kimi K2.5 ($-4.1$), and 61.2\% for MiniMax M2.5 ($-3.4$). These matches therefore account for only a limited share of successes; we report the stricter values as a conservative reference.

\begin{table}[htbp]
\centering
\captionsetup{skip=3pt}
\caption{\textbf{Self-consistency conditioned on a correct Definition.} Percentage of cases in which the model's Definition program agrees with its own answer in the given dimension, split by whether that answer is correct ($\checkmark$) or incorrect ($\times$). Denominators in parentheses. Pooled over both runs.}
\label{tab:selfcons_conditioned}
\footnotesize
\setlength{\tabcolsep}{3.4pt}
\renewcommand{\arraystretch}{0.95}
\resizebox{\columnwidth}{!}{%
\begin{tabular}{lcccccc}
\toprule
 & \multicolumn{2}{c}{\textbf{Classification}} & \multicolumn{2}{c}{\textbf{Constrained generation}} & \multicolumn{2}{c}{\textbf{Inversion}} \\
\cmidrule(lr){2-3}\cmidrule(lr){4-5}\cmidrule(lr){6-7}
\textbf{Model} & answer $\checkmark$ & answer $\times$ & answer $\checkmark$ & answer $\times$ & answer $\checkmark$ & answer $\times$ \\
\midrule
\multicolumn{7}{l}{\textsc{ARC-AGI-1 Train}}\\
GPT-5.4          & 99.3 {\scriptsize(3095)} & 14.4 {\scriptsize(90)}  & 97.0 {\scriptsize(526)} & 31.9 {\scriptsize(113)} & 98.2 {\scriptsize(559)} & 65.0 {\scriptsize(80)} \\
Gemini 3.1 Pro   & 99.3 {\scriptsize(3211)} & 3.2 {\scriptsize(154)}  & 99.4 {\scriptsize(622)} & 48.0 {\scriptsize(50)}  & 99.2 {\scriptsize(608)} & 73.0 {\scriptsize(63)} \\
Claude 4.6 Opus  & 99.2 {\scriptsize(2759)} & 4.1 {\scriptsize(196)}  & 98.4 {\scriptsize(491)} & 28.0 {\scriptsize(100)} & 98.5 {\scriptsize(545)} & 21.7 {\scriptsize(46)} \\
Kimi K2.5        & 99.6 {\scriptsize(2503)} & 8.9 {\scriptsize(202)}  & 98.4 {\scriptsize(446)} & 43.8 {\scriptsize(96)}  & 99.4 {\scriptsize(506)} & 35.1 {\scriptsize(37)} \\
MiniMax M2.5     & 99.3 {\scriptsize(2101)} & 3.1 {\scriptsize(419)}  & 96.7 {\scriptsize(391)} & 38.9 {\scriptsize(113)} & 97.4 {\scriptsize(431)} & 47.9 {\scriptsize(73)} \\
\specialrule{0.8pt}{0.4mm}{0.4mm}
\textbf{Pooled} & \multicolumn{6}{c}{\textbf{99.1} (answer $\checkmark$) \quad vs. \quad \textbf{21.1} (answer $\times$)} \\
\midrule
\multicolumn{7}{l}{\textsc{P-ARC}}\\
GPT-5.4          & 96.2 {\scriptsize(105)} & 40.0 {\scriptsize(5)}  & 100.0 {\scriptsize(13)} & 11.1 {\scriptsize(9)}  & 85.7 {\scriptsize(14)} & 25.0 {\scriptsize(8)} \\
Gemini 3.1 Pro   & 93.9 {\scriptsize(82)}  & 11.1 {\scriptsize(18)} & 100.0 {\scriptsize(15)} & 20.0 {\scriptsize(5)}  & 83.3 {\scriptsize(12)} & 25.0 {\scriptsize(8)} \\
Claude 4.6 Opus  & 93.9 {\scriptsize(230)} & 3.1 {\scriptsize(65)}  & 94.3 {\scriptsize(35)}  & 28.0 {\scriptsize(25)} & 93.3 {\scriptsize(45)} & 26.7 {\scriptsize(15)} \\
Kimi K2.5        & 93.7 {\scriptsize(63)}  & 23.5 {\scriptsize(17)} & 100.0 {\scriptsize(8)}  & 37.5 {\scriptsize(8)}  & 84.6 {\scriptsize(13)} & 0.0 {\scriptsize(3)} \\
MiniMax M2.5     & 94.3 {\scriptsize(35)}  & 20.0 {\scriptsize(15)} & 83.3 {\scriptsize(6)}   & 50.0 {\scriptsize(4)}  & 100.0 {\scriptsize(5)} & 20.0 {\scriptsize(5)} \\
\specialrule{0.8pt}{0.4mm}{0.4mm}
\textbf{Pooled} & \multicolumn{6}{c}{\textbf{94.0} (answer $\checkmark$) \quad vs. \quad \textbf{17.1} (answer $\times$)} \\
\bottomrule
\end{tabular}}
\end{table}

\subsection{Unconditional Self-Consistency}

\autoref{tab:self_consistency_combined} reports agreement between the Definition program and other dimensions without conditioning on Definition correctness. These rates are harder to interpret than the conditioned analysis in \autoref{sec:selfcons}: when the Definition is itself wrong, disagreement with it is uninformative about rule use. Classification agreement is substantially higher than for the two exact-generation dimensions on ARC-AGI-1, plausibly because Classification is a binary decision over externally provided candidates with many negatives, so agreement with the Definition-implied label is easier to achieve than exact grid reconstruction. On P-ARC agreement is lower and more balanced, consistent with less reliable Definition programs.

\begin{table}[htbp]
\centering
\captionsetup{skip=3pt}
\caption{\textbf{Unconditional self-consistency by dataset and model.} Panel (a) reports the percentage of cases in which the Definition program agrees with the model response in the corresponding dimension. Panel (b) restricts to model error cases.}
\label{tab:self_consistency_combined}
\footnotesize
\setlength{\tabcolsep}{3.8pt}
\renewcommand{\arraystretch}{0.92}
\resizebox{\columnwidth}{!}{%
\begin{tabular}{lcccccc}
\toprule
\multicolumn{7}{l}{\textbf{(a) All cases}}\\[-0.5mm]
        & \multicolumn{3}{c}{\textbf{ARC-AGI1}} & \multicolumn{3}{c}{\textbf{P-ARC}} \\
        \cmidrule(lr){2-4} \cmidrule(lr){5-7}
        \textbf{Model}
        & \makecell{\textbf{Classification}} & \makecell{\textbf{Constrained}\\ \textbf{generation}} & \makecell{\textbf{Inversion}}
        & \makecell{\textbf{Classification}} & \makecell{\textbf{Constrained}\\ \textbf{generation}} & \makecell{\textbf{Inversion}} \\
\midrule
Claude 4.6 Opus & 88.23 & 74.56 & 78.61 & 66.67 & 55.00 & 62.00 \\[-0.2mm]
GPT-5.4 & 93.62 & 77.01 & 83.17 & 68.37 & 37.76 & 31.63 \\[-0.2mm]
Gemini 3.1 Pro & 90.48 & 87.17 & 87.15 & 63.43 & 47.47 & 30.30 \\[-0.2mm]
Kimi K2.5 & 83.50 & 66.67 & 71.05 & 55.80 & 22.22 & 19.19 \\[-0.2mm]
MiniMax M2.5 & 72.94 & 61.35 & 66.28 & 48.27 & 22.67 & 10.81 \\
\specialrule{1.0pt}{0.5mm}{0.5mm}
\textbf{Aggregate} & \textbf{85.83} & \textbf{73.43} & \textbf{77.32} & \textbf{61.10} & \textbf{37.79} & \textbf{31.91} \\
\bottomrule
\end{tabular}}

\vspace{0.25em}

\resizebox{\columnwidth}{!}{%
\begin{tabular}{lcccccc}
\toprule
\multicolumn{7}{l}{\textbf{(b) Model mistakes only}}\\[-0.5mm]
        & \multicolumn{3}{c}{\textbf{ARC-AGI1}} & \multicolumn{3}{c}{\textbf{P-ARC}} \\
        \cmidrule(lr){2-4} \cmidrule(lr){5-7}
        \textbf{Model}
        & \makecell{\textbf{Classification}} & \makecell{\textbf{Constrained}\\ \textbf{generation}} & \makecell{\textbf{Inversion}}
        & \makecell{\textbf{Classification}} & \makecell{\textbf{Constrained}\\ \textbf{generation}} & \makecell{\textbf{Inversion}} \\
\midrule
Claude 4.6 Opus & 5.73 & 25.61 & 24.64 & 7.14 & 25.00 & 24.14 \\[-0.2mm]
GPT-5.4 & 27.98 & 29.28 & 52.10 & 26.88 & 23.33 & 23.53 \\[-0.2mm]
Gemini 3.1 Pro & 11.58 & 46.81 & 61.96 & 22.22 & 24.56 & 24.49 \\[-0.2mm]
Kimi K2.5 & 12.50 & 26.15 & 16.22 & 12.32 & 11.67 & 7.69 \\[-0.2mm]
MiniMax M2.5 & 5.81 & 23.51 & 28.25 & 10.08 & 13.79 & 1.85 \\
\specialrule{1.0pt}{0.5mm}{0.5mm}
\textbf{Aggregate} & \textbf{9.98} & \textbf{28.00} & \textbf{35.92} & \textbf{15.14} & \textbf{19.35} & \textbf{15.32} \\
\bottomrule
\end{tabular}}
\end{table}

\FloatBarrier

\subsection{Increased Sampling Budget}

\begin{table}[htbp]
\captionsetup{skip=5pt}
\centering
\caption{GPT-5.4 sampling budget trends over 20 ARC-AGI-1 tasks, each evaluated under 10 sampled configurations across all five dimensions (200 complete five-dimension evaluations). Entries are percentages, reported as cumulative mean $\pm$ sample standard deviation across sampled configurations.}
\label{tab:gpt54_subset20_sampling_budget_trends}
\footnotesize
\setlength{\tabcolsep}{5.6pt}
\renewcommand{\arraystretch}{1.05}
\begin{tabular}{lcccc}
\toprule
\makecell{\textbf{Dimension / Measure}} & \makecell{\textbf{Sample 1}} & \makecell{\textbf{Sample 2}} & \makecell{\textbf{Sample 5}} & \makecell{\textbf{Sample 10}} \\
\midrule
Avg. full-task acc. & 65.0 {\scriptsize$\pm$24.7} & 65.0 {\scriptsize$\pm$20.2} & 71.0 {\scriptsize$\pm$13.2} & 70.0 {\scriptsize$\pm$11.0} \\
Hard full-task acc. & 65.0 & 45.0 & 35.0 & 25.0 \\
Definition & 100.0 {\scriptsize$\pm$0.0} & 100.0 {\scriptsize$\pm$0.0} & 100.0 {\scriptsize$\pm$0.0} & 100.0 {\scriptsize$\pm$0.0} \\
Classification & 90.0 {\scriptsize$\pm$7.1} & 90.0 {\scriptsize$\pm$5.8} & 92.0 {\scriptsize$\pm$5.2} & 92.0 {\scriptsize$\pm$6.1} \\
Generation & 80.0 {\scriptsize$\pm$14.1} & 87.5 {\scriptsize$\pm$10.4} & 85.0 {\scriptsize$\pm$8.8} & 83.5 {\scriptsize$\pm$7.4} \\
Editing & 95.0 {\scriptsize$\pm$3.5} & 90.0 {\scriptsize$\pm$7.6} & 95.0 {\scriptsize$\pm$5.8} & 94.5 {\scriptsize$\pm$5.9} \\
Inversion & 90.0 {\scriptsize$\pm$7.1} & 90.0 {\scriptsize$\pm$5.8} & 89.0 {\scriptsize$\pm$4.9} & 91.0 {\scriptsize$\pm$4.0} \\
\bottomrule
\end{tabular}
\end{table}

\FloatBarrier

\begin{table}[htbp]
\centering
\caption{\textbf{Run-to-run stability of full-task pass sets.} For each model, the number of ARC-AGI-1 tasks passing all five dimensions in each run, their intersection and union, the Jaccard overlap, and the percentage of tasks on which the two runs agree (both pass or both fail). P-ARC is omitted: with 0--4 passing tasks per run, the overlap statistic is not meaningful.}
\label{tab:run_overlap}
\footnotesize
\begin{tabular}{lccccc}
\toprule
\textbf{Model} & \textbf{Run 1} & \textbf{Run 2} & \textbf{Shared} & \textbf{Jaccard} & \textbf{Task agreement} \\
\midrule
GPT-5.4 Low                & 185 & 188 & 146 & 0.64 & 79.8\% \\
Gemini 3.1 Pro Low         & 239 & 227 & 205 & 0.79 & 86.0\% \\
Claude 4.6 Opus Low (120K) & 160 & 141 & 107 & 0.55 & 78.2\% \\
Kimi K2.5                  & 162 & 147 & 102 & 0.49 & 73.8\% \\
MiniMax M2.5               &  76 &  89 &  57 & 0.53 & 87.2\% \\
\bottomrule
\end{tabular}
\end{table}

\begin{table}[htbp]
\centering
\caption{Failure rates by corruption type with pooled Wilson $95\%$ confidence intervals, over both runs. ``\textit{items per model}'' is the median candidate count for that type; counts vary by at most $1\%$ across models. Several P-ARC cells rest on few candidates (14 for color flip, 42 for retrieval) and the corresponding intervals are correspondingly wide.}
\label{tab:failure_rates_ci}
\scriptsize
\setlength{\tabcolsep}{3.2pt}
\resizebox{\columnwidth}{!}{
\begin{tabular}{lcccccc}
\toprule
\textbf{Model} & \textbf{Human} & \makecell{\textbf{Corrupted}\\\textbf{verifier}} & \textbf{Correct} & \makecell{\textbf{Color}\\\textbf{flip}} & \textbf{Retrieval} & \makecell{\textbf{Instance}\\\textbf{mismatch}} \\
\midrule
\multicolumn{7}{l}{\textsc{ARC-AGI-1}}\\
Claude 4.6 Opus & 20.74 \ci{17.6}{24.3} & 11.14 \ci{8.5}{14.5} & 1.84 \ci{1.2}{2.8} & 8.76 \ci{6.2}{12.2} & 4.52 \ci{2.9}{7.0} & 4.59 \ci{3.5}{6.0} \\
GPT-5.4 & 8.21 \ci{6.2}{10.8} & 2.30 \ci{1.3}{4.2} & 5.06 \ci{3.9}{6.5} & 6.53 \ci{4.4}{9.6} & 3.15 \ci{1.8}{5.3} & 1.85 \ci{1.2}{2.8} \\
Gemini 3.1 Pro & 16.37 \ci{13.5}{19.7} & 8.99 \ci{6.6}{12.0} & 5.48 \ci{4.3}{6.9} & 2.82 \ci{1.5}{5.1} & 2.17 \ci{1.1}{4.1} & 6.81 \ci{5.5}{8.5} \\
Kimi K2.5 & 18.54 \ci{15.5}{22.0} & 11.24 \ci{8.6}{14.6} & 5.98 \ci{4.7}{7.5} & 25.78 \ci{21.5}{30.6} & 10.79 \ci{8.2}{14.1} & 7.75 \ci{6.3}{9.5} \\
MiniMax M2.5 & 31.32 \ci{27.6}{35.3} & 26.62 \ci{22.7}{31.0} & 3.66 \ci{2.7}{4.9} & 40.23 \ci{35.2}{45.4} & 40.43 \ci{35.8}{45.2} & 18.12 \ci{15.9}{20.5} \\
\textit{items per model} & 562 & 432 & 1146 & 353 & 417 & 1086 \\
\midrule
\multicolumn{7}{l}{\textsc{P-ARC}}\\
Claude 4.6 Opus & 65.03 \ci{56.9}{72.4} & 12.50 \ci{6.5}{22.8} & 4.93 \ci{2.4}{9.8} & 28.57 \ci{11.7}{54.6} & 7.14 \ci{2.5}{19.0} & 12.22 \ci{7.0}{20.6} \\
GPT-5.4 & 47.22 \ci{39.2}{55.3} & 0.00 \ci{0.0}{5.7} & 16.67 \ci{11.5}{23.6} & 0.00 \ci{0.0}{21.5} & 0.00 \ci{0.0}{8.4} & 3.26 \ci{1.1}{9.2} \\
Gemini 3.1 Pro & 60.42 \ci{52.3}{68.0} & 9.38 \ci{4.4}{19.0} & 22.92 \ci{16.8}{30.4} & 14.29 \ci{4.0}{39.9} & 2.38 \ci{0.4}{12.3} & 7.61 \ci{3.7}{14.9} \\
Kimi K2.5 & 61.11 \ci{53.0}{68.7} & 14.06 \ci{7.6}{24.6} & 12.59 \ci{8.1}{19.0} & 42.86 \ci{21.4}{67.4} & 14.29 \ci{6.7}{27.8} & 11.83 \ci{6.7}{20.0} \\
MiniMax M2.5 & 70.83 \ci{62.9}{77.6} & 23.44 \ci{14.7}{35.1} & 12.50 \ci{8.1}{18.9} & 50.00 \ci{26.8}{73.2} & 30.95 \ci{19.1}{46.0} & 23.91 \ci{16.4}{33.6} \\
\textit{items per model} & 144 & 64 & 144 & 14 & 42 & 92 \\
\bottomrule
\end{tabular}}
\end{table}

\section{Definition Dimension Error Analysis}\label{sec:definition_errors}

Because the Definition dimension asks for executable code, some failures could in principle reflect programming or interface problems rather than incorrect rule inference. They largely do not. \autoref{tab:definition_exec} reports the share of collected programs that fail to load at all: between $0\%$ and $2\%$ for every model except MiniMax M2.5, whose $3.6\%$ (ARC-AGI-1) and $25\%$ (P-ARC) rates stem mainly from mangled code or natural-language text mixed into the program, apparently a side effect of substantially longer thinking chains.

We classify each Definition failure as \emph{static} (the program does not load), \emph{runtime} (it loads, but every failure across the four scored splits is a raised exception), or \emph{rule} (it loads and returns at least one incorrect output grid). A program that returns a wrong grid on any scored split counts as a rule failure even if it also raises an exception elsewhere. Pooled over both runs and all five models, this gives a static/runtime/rule split of $4\%/12\%/84\%$ on ARC-AGI-1 and $6\%/9\%/85\%$ on P-ARC; excluding MiniMax M2.5, which dominates the static counts, $3\%/11\%/87\%$ and $1\%/6\%/93\%$. Runtime failures are concentrated in the two open-weight models: they account for $22.5\%$ of Kimi K2.5's and $16.1\%$ of MiniMax's ARC-AGI-1 Definition failures, against $1.9\%$ for GPT-5.4 and $2.4\%$ for Gemini 3.1 Pro. We do not separate overfitting to the demonstrations from incorrect rule inference, as the former is a special case of the latter. Overall, most Definition failures come from formalizing the wrong rule, not from producing non-executable code.

\begin{table}[htbp]
\centering
\caption{Share of collected Definition programs that fail to load.}
\label{tab:definition_exec}
\footnotesize
\begin{tabular}{llccc}
\toprule
\textbf{Model} & \textbf{Suite} & \textbf{Programs} & \textbf{Failed to load} & \textbf{\%} \\
\midrule
GPT-5.4 & ARC-AGI-1 & 800 & 4 & 0.5 \\
Gemini 3.1 Pro & ARC-AGI-1 & 800 & 4 & 0.5 \\
Claude 4.6 Opus & ARC-AGI-1 & 800 & 10 & 1.2 \\
Kimi K2.5 & ARC-AGI-1 & 800 & 8 & 1.0 \\
MiniMax M2.5 & ARC-AGI-1 & 800 & 29 & 3.6 \\
\midrule
GPT-5.4 & P-ARC & 100 & 2 & 2.0 \\
Gemini 3.1 Pro & P-ARC & 100 & 1 & 1.0 \\
Claude 4.6 Opus & P-ARC & 100 & 0 & 0.0 \\
Kimi K2.5 & P-ARC & 100 & 0 & 0.0 \\
MiniMax M2.5 & P-ARC & 100 & 25 & 25.0 \\
\bottomrule
\end{tabular}
\end{table}

\section{P-ARC Construction and Calibration}\label{sec:parc_details}

\textbf{Taxonomy.} Grouping the 50 tasks by their primary underlying operation gives: physics and dynamics, such as gravity, motion, bouncing and flow (9 tasks); rigid geometric transforms such as rotation, reflection, scaling and wrapping (8); connection and pathfinding, including shape connection, routing and minimum spanning trees (7); symmetry, completion, occlusion recovery and frame repair (8); object composition and assembly (5); constraint satisfaction and logic, such as sudoku completion, map coloring and XOR of halves (6); and color or attribute mapping, such as color shifts and directional shadowing (5). Two tasks have no clean category assignment. Per-task rule descriptions and category assignments are released with the dataset.

\textbf{Feasibility check.} Our human protocol is a feasibility check embedded in the design process, not a formal human study with naive participants. Each candidate task was shown to team members other than its author---at least one, most often two or three---and retained only if at least one produced the correct output. Multiple attempts were permitted, but tasks were typically solved within one or two guesses. Solve times were not recorded, so we report no solve-rate or solve-time statistics; a formal human study is left to future work, and our positioning of P-ARC between ARC-AGI-1 and ARC-AGI-2 in difficulty remains a qualitative estimate. Erroneous attempts collected during this process provide part of the human corruptions; the remaining ones were designed by each task's creator, giving three incorrect output grids per task.

\textbf{Generator and verifier validation.} For every task, a team member other than the task's author manually inspected at least 50 generator-produced examples and confirmed that they realize the intended rule; tasks were admitted only once this held. This validates generator and verifier behavior on their sampled outputs, which is the property our evaluation depends on, and complements the generator/verifier alignment limitation discussed in the main text.

\textbf{Release.} P-ARC is released under the MIT license at \url{https://huggingface.co/datasets/ClaasBeger/P-ARC}. Each task ships in the standard ARC JSON schema together with its generator and verifier programs, 50 stable generated examples, the three human-made corruptions, and a natural-language statement of its rule.

\section{Thinking Token Consumption by Claude Opus 4.6}\label{sec:adaptive_token}

In our experiments, Claude Opus 4.6 shows a clear increase in thinking-token consumption when moving from ARC-AGI-1 to P-ARC. After accounting for pricing differences, its token consumption on P-ARC is roughly twice that of Gemini or GPT, whereas consumption is comparable across models on ARC-AGI-1. Prompt-to-response latency increases even more sharply, yielding approximately threefold longer evaluation time, with extreme outliers requiring up to an hour for a single dimension. This initially constrained the evaluation budget, but we subsequently completed a second inference pass so that the final reported results use two runs.

We also evaluate Claude without a thinking budget in one full additional run on the ARC-AGI-1 training set. Relative to the pooled two-run configuration reported in \autoref{tab:main_results}, full-task accuracy drops from $37.6\%$ to $21.8\%$. Dimension-level performance decreases across the board: Definition $66.8 \to 48.0$ ($-18.8$ points), Classification $71.5 \to 62.0$ ($-9.5$), Constrained Generation $78.9 \to 58.0$ ($-20.9$), Editing $88.2 \to 65.5$ ($-22.7$), and Inversion $91.0 \to 75.8$ ($-15.2$). The largest drop is in Editing and the smallest in Classification, indicating that enabling thinking is an important driver of Claude's performance on PotARCin.

\section{Compute Budget and Hyperparameters}
For all API requests, we set the temperature to 1.0 whenever this option was available. For most proprietary models evaluated, this matches the default setting when reasoning or thinking mode is enabled. For GPT-5.4 with reasoning enabled, including the generative-sampling study, the API does not expose a user-configurable temperature parameter. The embeddings used for the retrieval corruption type were computed on a MacBook Pro with an Apple M4 Pro chip. Overall, our experiments do not rely on compute-intensive procedures or hardware-specific optimizations that would materially affect reproducibility. The random seed used for all reported runs is 77 and is recorded in the run logs.

\clearpage

\section{Dimension Prompts}
\subsection{Definition Prompt \label{app:definition_prompt}}
\begin{lessonbox}[top=0.6ex]
Find the common rule that maps an input grid to an output grid, given the examples below. Generate a python script that takes a grid as input and transforms it according to this rule. Anticipate that this program will be applied to several grids, not only the shown test grid.

\medskip
Your script must define exactly one callable entrypoint:
\begin{gridbox}
def solve(grid):
\end{gridbox}
where \texttt{grid} is passed as a list of lists of integers (\texttt{list[list[int]]}) and the function returns the transformed grid in the same format.

\medskip
\textbf{Example 1}

\emph{Input:}
\begin{gridbox}
<TRAIN_INPUT_1_GRID>
\end{gridbox}

\emph{Output:}
\begin{gridbox}
<TRAIN_OUTPUT_1_GRID>
\end{gridbox}

\medskip
\textbf{Example 2}

\emph{Input:}
\begin{gridbox}
<TRAIN_INPUT_2_GRID>
\end{gridbox}

\emph{Output:}
\begin{gridbox}
<TRAIN_OUTPUT_2_GRID>
\end{gridbox}

\medskip
\emph{\textit{... one block per training example}}

\medskip
Below is a test input grid.

\medskip
\emph{Input:}
\begin{gridbox}
<TEST_INPUT_GRID>
\end{gridbox}

\medskip
\textbf{Your final answer should just be the python script itself, no other text or markdown.}
\end{lessonbox}

\subsection{Classification Distribution Prompt \label{app:classification_prompt}}
\begin{lessonbox}[top=0.6ex]
Find the common rule that maps an input grid to an output grid given the examples below. You will see five additional input/output pairs (test items 0--4). For each test item, decide whether that pair could have been produced by the same underlying rule and data distribution as the training examples (1 = same distribution / consistent, 0 = not).

\medskip
\textbf{Training example 1}

\emph{Input:}
\begin{gridbox}
<TRAIN_INPUT_1_GRID>
\end{gridbox}

\emph{Output:}
\begin{gridbox}
<TRAIN_OUTPUT_1_GRID>
\end{gridbox}

\medskip
\textbf{Training example 2}

\emph{Input:}
\begin{gridbox}
<TRAIN_INPUT_2_GRID>
\end{gridbox}

\emph{Output:}
\begin{gridbox}
<TRAIN_OUTPUT_2_GRID>
\end{gridbox}

\medskip
\emph{\textit{... one block per training example}}

\medskip
Test pairs (evaluate each in order, index 0 \ldots{} 4):

\medskip
\textbf{Test 0}

\emph{Input:}
\begin{gridbox}
<TEST0_INPUT_GRID>
\end{gridbox}

\emph{Output:}
\begin{gridbox}
<TEST0_OUTPUT_GRID>
\end{gridbox}

\medskip
\textbf{Test 1}

\emph{Input:}
\begin{gridbox}
<TEST1_INPUT_GRID>
\end{gridbox}

\emph{Output:}
\begin{gridbox}
<TEST1_OUTPUT_GRID>
\end{gridbox}

\medskip
\emph{\textit{... through Test 4}}

\medskip
\textbf{Your final answer must be only a single Python list of exactly five integers, each 0 or 1, in order for test 0 through test 4, for example [1, 0, 1, 1, 0]. Output nothing else: no other text, no markdown code fences, and no explanation.}
\end{lessonbox}

\subsection{Constrained Generation Prompt \label{app:constrained_generation_prompt}}
\begin{lessonbox}[top=0.6ex]
Find the common rule that maps an input grid to an output grid given the examples below.

\medskip
\textbf{Example 1}

\emph{Input:}
\begin{gridbox}
<TRAIN_INPUT_1_GRID>
\end{gridbox}

\emph{Output:}
\begin{gridbox}
<TRAIN_OUTPUT_1_GRID>
\end{gridbox}

\medskip
\textbf{Example 2}

\emph{Input:}
\begin{gridbox}
<TRAIN_INPUT_2_GRID>
\end{gridbox}

\emph{Output:}
\begin{gridbox}
<TRAIN_OUTPUT_2_GRID>
\end{gridbox}

\medskip
\emph{\textit{... one block per training example}}

\medskip
Generate one NEW input/output pair that follows the same transformation rule.
\begin{itemize}[leftmargin=*, itemsep=2pt, topsep=2pt]
    \item Do NOT copy any demonstration pair exactly.
    \item The generated input and output must be different (non-trivial transformation).
    \item Your pair must be consistent with the same underlying rule.
\end{itemize}

\medskip
\textbf{Your final answer must be ONLY a single Python dict with exactly these keys:}
\begin{gridbox}
{"input": [[...], ...], "output": [[...], ...]}
\end{gridbox}
Use integer grid cells 0--9. Output nothing else.
\end{lessonbox}

\clearpage

\subsection{Editing Prompt \label{app:editing_prompt}}
\begin{lessonbox}[top=0.6ex]
Find the common rule that maps an input grid to an output grid given the examples below.

\medskip
\textbf{Example 1}

\emph{Input:}
\begin{gridbox}
<TRAIN_INPUT_1_GRID>
\end{gridbox}

\emph{Output:}
\begin{gridbox}
<TRAIN_OUTPUT_1_GRID>
\end{gridbox}

\medskip
\textbf{Example 2}

\emph{Input:}
\begin{gridbox}
<TRAIN_INPUT_2_GRID>
\end{gridbox}

\emph{Output:}
\begin{gridbox}
<TRAIN_OUTPUT_2_GRID>
\end{gridbox}

\medskip
\emph{\textit{... one block per training example}}

\medskip
\begin{tcolorbox}[
  colback=white, colframe=white, boxrule=0pt, boxsep=0pt,
  left=0pt, right=0pt, top=0pt, bottom=0pt]
The following candidate input/output pair has been corrupted in some way. Your task is to return the corrected input/output pair that is consistent with the rule inferred from the training examples.
\end{tcolorbox}

\medskip
\textbf{Candidate pair}

\emph{Input:}
\begin{gridbox}
<CANDIDATE_INPUT_GRID>
\end{gridbox}

\emph{Output:}
\begin{gridbox}
<CANDIDATE_OUTPUT_GRID>
\end{gridbox}

\medskip
\textbf{Your final answer must be ONLY a single Python dict with exactly these keys:}
\begin{gridbox}
{"input": [[...], ...], "output": [[...], ...]}
\end{gridbox}
Use integer grid cells 0--9. Output nothing else.
\end{lessonbox}

\subsection{Inversion Prompt \label{app:inversion_prompt}}
\begin{lessonbox}[top=0.6ex]
Find the common rule that maps an input grid to an output grid given the examples below.

\medskip
\textbf{Example 1}

\emph{Input:}
\begin{gridbox}
<TRAIN_INPUT_1_GRID>
\end{gridbox}

\emph{Output:}
\begin{gridbox}
<TRAIN_OUTPUT_1_GRID>
\end{gridbox}

\medskip
\textbf{Example 2}

\emph{Input:}
\begin{gridbox}
<TRAIN_INPUT_2_GRID>
\end{gridbox}

\emph{Output:}
\begin{gridbox}
<TRAIN_OUTPUT_2_GRID>
\end{gridbox}

\medskip
\emph{\textit{... one block per training example}}

\medskip
Now infer an input grid that would produce the following output by applying the same rule.

\medskip
\emph{Desired output:}
\begin{gridbox}
<DESIRED_OUTPUT_GRID>
\end{gridbox}

\medskip
\textbf{Your final answer must be ONLY a single Python list-of-lists grid:}
\begin{gridbox}
[[...], ...]
\end{gridbox}
Use integer grid cells 0--9. Output nothing else.
\end{lessonbox}

\subsection{Output Grid Correctness Prompt \label{app:output_grid_prompt}}
\begin{lessonbox}[top=0.6ex]
Find the common rule that maps an input grid to an output grid, given the examples below.

\medskip
\textbf{Example 1}

\emph{Input:}
\begin{gridbox}
<TRAIN_INPUT_1_GRID>
\end{gridbox}

\emph{Output:}
\begin{gridbox}
<TRAIN_OUTPUT_1_GRID>
\end{gridbox}

\medskip
\textbf{Example 2}

\emph{Input:}
\begin{gridbox}
<TRAIN_INPUT_2_GRID>
\end{gridbox}

\emph{Output:}
\begin{gridbox}
<TRAIN_OUTPUT_2_GRID>
\end{gridbox}

\medskip
\emph{\textit{... one block per training example}}

\medskip
Below is a test input grid. Predict the corresponding output grid by applying the rule you found.

\medskip
\emph{Test Input:}
\begin{gridbox}
<TEST_INPUT_GRID>
\end{gridbox}

\medskip
\textbf{Your final answer must be ONLY a single Python list-of-lists grid for that output:}
\begin{gridbox}
[[...], ...]
\end{gridbox}
Use integer grid cells 0--9. Output nothing else.
\end{lessonbox}
\newpage
\section*{NeurIPS Paper Checklist}

\begin{enumerate}

\item {\bf Claims}
    \item[] Question: Do the main claims made in the abstract and introduction accurately reflect the paper's contributions and scope?
    \item[] Answer: \answerYes{} 
    \item[] Justification: The abstract and introduction describe PotARCin, the five evaluation dimensions, P-ARC, and the main empirical finding that performance drops substantially relative to standard ARC output-grid accuracy. These claims are supported by the methodology and results sections.
    \item[] Guidelines:
    \begin{itemize}
        \item The answer \answerNA{} means that the abstract and introduction do not include the claims made in the paper.
        \item The abstract and/or introduction should clearly state the claims made, including the contributions made in the paper and important assumptions and limitations. A \answerNo{} or \answerNA{} answer to this question will not be perceived well by the reviewers. 
        \item The claims made should match theoretical and experimental results, and reflect how much the results can be expected to generalize to other settings. 
        \item It is fine to include aspirational goals as motivation as long as it is clear that these goals are not attained by the paper. 
    \end{itemize}

\item {\bf Limitations}
    \item[] Question: Does the paper discuss the limitations of the work performed by the authors?
    \item[] Answer: \answerYes{} 
    \item[] Justification: The paper includes a Limitations and Future Work paragraph discussing dependence on generator/verifier alignment, model-compute comparability, the mimetic similarity restriction, and the limited treatment of example diversity and expressiveness.
    \item[] Guidelines:
    \begin{itemize}
        \item The answer \answerNA{} means that the paper has no limitation while the answer \answerNo{} means that the paper has limitations, but those are not discussed in the paper. 
        \item The authors are encouraged to create a separate ``Limitations'' section in their paper.
        \item The paper should point out any strong assumptions and how robust the results are to violations of these assumptions (e.g., independence assumptions, noiseless settings, model well-specification, asymptotic approximations only holding locally). The authors should reflect on how these assumptions might be violated in practice and what the implications would be.
        \item The authors should reflect on the scope of the claims made, e.g., if the approach was only tested on a few datasets or with a few runs. In general, empirical results often depend on implicit assumptions, which should be articulated.
        \item The authors should reflect on the factors that influence the performance of the approach. For example, a facial recognition algorithm may perform poorly when image resolution is low or images are taken in low lighting. Or a speech-to-text system might not be used reliably to provide closed captions for online lectures because it fails to handle technical jargon.
        \item The authors should discuss the computational efficiency of the proposed algorithms and how they scale with dataset size.
        \item If applicable, the authors should discuss possible limitations of their approach to address problems of privacy and fairness.
        \item While the authors might fear that complete honesty about limitations might be used by reviewers as grounds for rejection, a worse outcome might be that reviewers discover limitations that aren't acknowledged in the paper. The authors should use their best judgment and recognize that individual actions in favor of transparency play an important role in developing norms that preserve the integrity of the community. Reviewers will be specifically instructed to not penalize honesty concerning limitations.
    \end{itemize}

\item {\bf Theory assumptions and proofs}
    \item[] Question: For each theoretical result, does the paper provide the full set of assumptions and a complete (and correct) proof?
    \item[] Answer: \answerNA{} 
    \item[] Justification: The paper does not present new theoretical results or formal proofs; its contributions are empirical and benchmark-oriented.
    \item[] Guidelines:
    \begin{itemize}
        \item The answer \answerNA{} means that the paper does not include theoretical results. 
        \item All the theorems, formulas, and proofs in the paper should be numbered and cross-referenced.
        \item All assumptions should be clearly stated or referenced in the statement of any theorems.
        \item The proofs can either appear in the main paper or the supplemental material, but if they appear in the supplemental material, the authors are encouraged to provide a short proof sketch to provide intuition. 
        \item Inversely, any informal proof provided in the core of the paper should be complemented by formal proofs provided in appendix or supplemental material.
        \item Theorems and Lemmas that the proof relies upon should be properly referenced. 
    \end{itemize}

    \item {\bf Experimental result reproducibility}
    \item[] Question: Does the paper fully disclose all the information needed to reproduce the main experimental results of the paper to the extent that it affects the main claims and/or conclusions of the paper (regardless of whether the code and data are provided or not)?
    \item[] Answer: \answerYes{} 
    \item[] Justification: The methodology describes the datasets, generator and verifier construction, evaluation dimensions, corruption types, model set, sampling protocol, and evaluation metrics, and the appendix gives further implementation details. Code, model outputs and the exact run commands are available at \url{https://github.com/ClaasBeger/PotARCin}, and P-ARC at \url{https://huggingface.co/datasets/ClaasBeger/P-ARC}.
    \item[] Guidelines:
    \begin{itemize}
        \item The answer \answerNA{} means that the paper does not include experiments.
        \item If the paper includes experiments, a \answerNo{} answer to this question will not be perceived well by the reviewers: Making the paper reproducible is important, regardless of whether the code and data are provided or not.
        \item If the contribution is a dataset and\slash or model, the authors should describe the steps taken to make their results reproducible or verifiable. 
        \item Depending on the contribution, reproducibility can be accomplished in various ways. For example, if the contribution is a novel architecture, describing the architecture fully might suffice, or if the contribution is a specific model and empirical evaluation, it may be necessary to either make it possible for others to replicate the model with the same dataset, or provide access to the model. In general. releasing code and data is often one good way to accomplish this, but reproducibility can also be provided via detailed instructions for how to replicate the results, access to a hosted model (e.g., in the case of a large language model), releasing of a model checkpoint, or other means that are appropriate to the research performed.
        \item While NeurIPS does not require releasing code, the conference does require all submissions to provide some reasonable avenue for reproducibility, which may depend on the nature of the contribution. For example
        \begin{enumerate}
            \item If the contribution is primarily a new algorithm, the paper should make it clear how to reproduce that algorithm.
            \item If the contribution is primarily a new model architecture, the paper should describe the architecture clearly and fully.
            \item If the contribution is a new model (e.g., a large language model), then there should either be a way to access this model for reproducing the results or a way to reproduce the model (e.g., with an open-source dataset or instructions for how to construct the dataset).
            \item We recognize that reproducibility may be tricky in some cases, in which case authors are welcome to describe the particular way they provide for reproducibility. In the case of closed-source models, it may be that access to the model is limited in some way (e.g., to registered users), but it should be possible for other researchers to have some path to reproducing or verifying the results.
        \end{enumerate}
    \end{itemize}

\item {\bf Open access to data and code}
    \item[] Question: Does the paper provide open access to the data and code, with sufficient instructions to faithfully reproduce the main experimental results, as described in supplemental material?
    \item[] Answer: \answerYes{} 
    \item[] Justification: We release the evaluation code, prompts, P-ARC generators and verifiers, and all model outputs and scores at \url{https://github.com/ClaasBeger/PotARCin} under the MIT license; its README gives the setup and the exact command for each evaluation. P-ARC is also available at \url{https://huggingface.co/datasets/ClaasBeger/P-ARC}.
    \item[] Guidelines:
    \begin{itemize}
        \item The answer \answerNA{} means that paper does not include experiments requiring code.
        \item Please see the NeurIPS code and data submission guidelines (\url{https://neurips.cc/public/guides/CodeSubmissionPolicy}) for more details.
        \item While we encourage the release of code and data, we understand that this might not be possible, so \answerNo{} is an acceptable answer. Papers cannot be rejected simply for not including code, unless this is central to the contribution (e.g., for a new open-source benchmark).
        \item The instructions should contain the exact command and environment needed to run to reproduce the results. See the NeurIPS code and data submission guidelines (\url{https://neurips.cc/public/guides/CodeSubmissionPolicy}) for more details.
        \item The authors should provide instructions on data access and preparation, including how to access the raw data, preprocessed data, intermediate data, and generated data, etc.
        \item The authors should provide scripts to reproduce all experimental results for the new proposed method and baselines. If only a subset of experiments are reproducible, they should state which ones are omitted from the script and why.
        \item At submission time, to preserve anonymity, the authors should release anonymized versions (if applicable).
        \item Providing as much information as possible in supplemental material (appended to the paper) is recommended, but including URLs to data and code is permitted.
    \end{itemize}

\item {\bf Experimental setting/details}
    \item[] Question: Does the paper specify all the training and test details (e.g., data splits, hyperparameters, how they were chosen, type of optimizer) necessary to understand the results?
    \item[] Answer: \answerYes{} 
    \item[] Justification: The paper specifies the evaluated datasets, task splits, models, number of runs, reasoning-effort settings, evaluation dimensions, full-task accuracy metric, sampling procedures, and corruption types. No model training or optimizer is introduced by this work.
    \item[] Guidelines:
    \begin{itemize}
        \item The answer \answerNA{} means that the paper does not include experiments.
        \item The experimental setting should be presented in the core of the paper to a level of detail that is necessary to appreciate the results and make sense of them.
        \item The full details can be provided either with the code, in appendix, or as supplemental material.
    \end{itemize}

\item {\bf Experiment statistical significance}
    \item[] Question: Does the paper report error bars suitably and correctly defined or other appropriate information about the statistical significance of the experiments?
    \item[] Answer: \answerYes{} 
    \item[] Justification: The paper includes standard deviations based on two runs for all reported evaluations.
    \item[] Guidelines:
    \begin{itemize}
        \item The answer \answerNA{} means that the paper does not include experiments.
        \item The authors should answer \answerYes{} if the results are accompanied by error bars, confidence intervals, or statistical significance tests, at least for the experiments that support the main claims of the paper.
        \item The factors of variability that the error bars are capturing should be clearly stated (for example, train/test split, initialization, random drawing of some parameter, or overall run with given experimental conditions).
        \item The method for calculating the error bars should be explained (closed form formula, call to a library function, bootstrap, etc.)
        \item The assumptions made should be given (e.g., Normally distributed errors).
        \item It should be clear whether the error bar is the standard deviation or the standard error of the mean.
        \item It is OK to report 1-sigma error bars, but one should state it. The authors should preferably report a 2-sigma error bar than state that they have a 96\% CI, if the hypothesis of Normality of errors is not verified.
        \item For asymmetric distributions, the authors should be careful not to show in tables or figures symmetric error bars that would yield results that are out of range (e.g., negative error rates).
        \item If error bars are reported in tables or plots, the authors should explain in the text how they were calculated and reference the corresponding figures or tables in the text.
    \end{itemize}

\item {\bf Experiments compute resources}
    \item[] Question: For each experiment, does the paper provide sufficient information on the computer resources (type of compute workers, memory, time of execution) needed to reproduce the experiments?
    \item[] Answer: \answerNo{} 
    \item[] Justification: The paper reports the evaluated models, reasoning-effort settings, number of runs, and token-consumption effects, but most experiments were conducted through cloud-based model APIs. As a result, we cannot report underlying compute-worker types, memory, or wall-clock execution details for every experiment.
    \item[] Guidelines:
    \begin{itemize}
        \item The answer \answerNA{} means that the paper does not include experiments.
        \item The paper should indicate the type of compute workers CPU or GPU, internal cluster, or cloud provider, including relevant memory and storage.
        \item The paper should provide the amount of compute required for each of the individual experimental runs as well as estimate the total compute. 
        \item The paper should disclose whether the full research project required more compute than the experiments reported in the paper (e.g., preliminary or failed experiments that didn't make it into the paper). 
    \end{itemize}
    
\item {\bf Code of ethics}
    \item[] Question: Does the research conducted in the paper conform, in every respect, with the NeurIPS Code of Ethics \url{https://neurips.cc/public/EthicsGuidelines}?
    \item[] Answer: \answerYes{} 
    \item[] Justification: The reported research fully adheres to the outlined Code of Ethics.
    \item[] Guidelines:
    \begin{itemize}
        \item The answer \answerNA{} means that the authors have not reviewed the NeurIPS Code of Ethics.
        \item If the authors answer \answerNo, they should explain the special circumstances that require a deviation from the Code of Ethics.
        \item The authors should make sure to preserve anonymity (e.g., if there is a special consideration due to laws or regulations in their jurisdiction).
    \end{itemize}

\item {\bf Broader impacts}
    \item[] Question: Does the paper discuss both potential positive societal impacts and negative societal impacts of the work performed?
    \item[] Answer: \answerNA{} 
    \item[] Justification: The current draft focuses on benchmark design and model evaluation on abstract reasoning tasks, and we do not expect a large societal impact.
    \item[] Guidelines:
    \begin{itemize}
        \item The answer \answerNA{} means that there is no societal impact of the work performed.
        \item If the authors answer \answerNA{} or \answerNo, they should explain why their work has no societal impact or why the paper does not address societal impact.
        \item Examples of negative societal impacts include potential malicious or unintended uses (e.g., disinformation, generating fake profiles, surveillance), fairness considerations (e.g., deployment of technologies that could make decisions that unfairly impact specific groups), privacy considerations, and security considerations.
        \item The conference expects that many papers will be foundational research and not tied to particular applications, let alone deployments. However, if there is a direct path to any negative applications, the authors should point it out. For example, it is legitimate to point out that an improvement in the quality of generative models could be used to generate Deepfakes for disinformation. On the other hand, it is not needed to point out that a generic algorithm for optimizing neural networks could enable people to train models that generate Deepfakes faster.
        \item The authors should consider possible harms that could arise when the technology is being used as intended and functioning correctly, harms that could arise when the technology is being used as intended but gives incorrect results, and harms following from (intentional or unintentional) misuse of the technology.
        \item If there are negative societal impacts, the authors could also discuss possible mitigation strategies (e.g., gated release of models, providing defenses in addition to attacks, mechanisms for monitoring misuse, mechanisms to monitor how a system learns from feedback over time, improving the efficiency and accessibility of ML).
    \end{itemize}
    
\item {\bf Safeguards}
    \item[] Question: Does the paper describe safeguards that have been put in place for responsible release of data or models that have a high risk for misuse (e.g., pre-trained language models, image generators, or scraped datasets)?
    \item[] Answer: \answerNA{} 
    \item[] Justification: The paper does not release a pretrained model, image generator, scraped dataset, or other asset with a high risk of misuse. The released assets are ARC-style grid tasks, generators, verifiers, and evaluation materials, which do not carry such a risk.
    \item[] Guidelines:
    \begin{itemize}
        \item The answer \answerNA{} means that the paper poses no such risks.
        \item Released models that have a high risk for misuse or dual-use should be released with necessary safeguards to allow for controlled use of the model, for example by requiring that users adhere to usage guidelines or restrictions to access the model or implementing safety filters. 
        \item Datasets that have been scraped from the Internet could pose safety risks. The authors should describe how they avoided releasing unsafe images.
        \item We recognize that providing effective safeguards is challenging, and many papers do not require this, but we encourage authors to take this into account and make a best faith effort.
    \end{itemize}

\item {\bf Licenses for existing assets}
    \item[] Question: Are the creators or original owners of assets (e.g., code, data, models), used in the paper, properly credited and are the license and terms of use explicitly mentioned and properly respected?
    \item[] Answer: \answerYes{} 
    \item[] Justification: We cite all existing assets and list each with its source and license in the repository README: ARC-AGI-1 (Apache 2.0), RE-ARC (MIT), ARC-GEN and its ARC-GEN-100K examples (Apache 2.0), H-ARC (CC0 1.0), V-ARC (MIT), and the code-golf verifier solutions (Apache 2.0). We modified the ARC-GEN generators of 110 tasks so that they reproduce each task's demonstrations and test pairs; the modified files are marked and listed, as the Apache license requires.
    \item[] Guidelines:
    \begin{itemize}
        \item The answer \answerNA{} means that the paper does not use existing assets.
        \item The authors should cite the original paper that produced the code package or dataset.
        \item The authors should state which version of the asset is used and, if possible, include a URL.
        \item The name of the license (e.g., CC-BY 4.0) should be included for each asset.
        \item For scraped data from a particular source (e.g., website), the copyright and terms of service of that source should be provided.
        \item If assets are released, the license, copyright information, and terms of use in the package should be provided. For popular datasets, \url{paperswithcode.com/datasets} has curated licenses for some datasets. Their licensing guide can help determine the license of a dataset.
        \item For existing datasets that are re-packaged, both the original license and the license of the derived asset (if it has changed) should be provided.
        \item If this information is not available online, the authors are encouraged to reach out to the asset's creators.
    \end{itemize}

\item {\bf New assets}
    \item[] Question: Are new assets introduced in the paper well documented and is the documentation provided alongside the assets?
    \item[] Answer: \answerYes{} 
    \item[] Justification: The paper introduces PotARCin and P-ARC, including 50 hand-crafted ARC-style tasks with generator and verifier programs, and the methodology describes their construction. The code is documented in the repository README and released under the MIT license. P-ARC is available at \url{https://huggingface.co/datasets/ClaasBeger/P-ARC} under the MIT license, with a dataset card documenting its contents, construction and columns.
    \item[] Guidelines:
    \begin{itemize}
        \item The answer \answerNA{} means that the paper does not release new assets.
        \item Researchers should communicate the details of the dataset\slash code\slash model as part of their submissions via structured templates. This includes details about training, license, limitations, etc. 
        \item The paper should discuss whether and how consent was obtained from people whose asset is used.
        \item At submission time, remember to anonymize your assets (if applicable). You can either create an anonymized URL or include an anonymized zip file.
    \end{itemize}

\item {\bf Crowdsourcing and research with human subjects}
    \item[] Question: For crowdsourcing experiments and research with human subjects, does the paper include the full text of instructions given to participants and screenshots, if applicable, as well as details about compensation (if any)? 
    \item[] Answer: \answerNA{} 
    \item[] Justification: The paper does not conduct a crowdsourcing experiment or human-subjects study. P-ARC grids and erroneous outputs were produced internally by members of the task-design team as part of benchmark construction, not collected from recruited participants.
    \item[] Guidelines:
    \begin{itemize}
        \item The answer \answerNA{} means that the paper does not involve crowdsourcing nor research with human subjects.
        \item Including this information in the supplemental material is fine, but if the main contribution of the paper involves human subjects, then as much detail as possible should be included in the main paper. 
        \item According to the NeurIPS Code of Ethics, workers involved in data collection, curation, or other labor should be paid at least the minimum wage in the country of the data collector. 
    \end{itemize}

\item {\bf Institutional review board (IRB) approvals or equivalent for research with human subjects}
    \item[] Question: Does the paper describe potential risks incurred by study participants, whether such risks were disclosed to the subjects, and whether Institutional Review Board (IRB) approvals (or an equivalent approval/review based on the requirements of your country or institution) were obtained?
    \item[] Answer: \answerNA{} 
    \item[] Justification: The work does not involve recruited human subjects. The human-generated P-ARC grids and corruptions were produced internally by the task-design team during dataset construction.
    \item[] Guidelines:
    \begin{itemize}
        \item The answer \answerNA{} means that the paper does not involve crowdsourcing nor research with human subjects.
        \item Depending on the country in which research is conducted, IRB approval (or equivalent) may be required for any human subjects research. If you obtained IRB approval, you should clearly state this in the paper. 
        \item We recognize that the procedures for this may vary significantly between institutions and locations, and we expect authors to adhere to the NeurIPS Code of Ethics and the guidelines for their institution. 
        \item For initial submissions, do not include any information that would break anonymity (if applicable), such as the institution conducting the review.
    \end{itemize}

\item {\bf Declaration of LLM usage}
    \item[] Question: Does the paper describe the usage of LLMs if it is an important, original, or non-standard component of the core methods in this research? Note that if the LLM is used only for writing, editing, or formatting purposes and does \emph{not} impact the core methodology, scientific rigor, or originality of the research, declaration is not required.
    \item[] Answer: \answerYes{} 
    \item[] Justification: LLMs are central to the experiments: the paper evaluates several frontier and open-source models. The methodology and results sections describe the models and evaluation settings.
    \item[] Guidelines:
    \begin{itemize}
        \item The answer \answerNA{} means that the core method development in this research does not involve LLMs as any important, original, or non-standard components.
        \item Please refer to our LLM policy in the NeurIPS handbook for what should or should not be described.
    \end{itemize}

\end{enumerate}

\end{document}